\documentclass[11pt]{article}
\usepackage{epsfig, epsf, graphicx, subfigure, color}
\usepackage{pstricks, pst-node, psfrag}
\usepackage{amssymb,amsmath}
\usepackage{verbatim,enumerate}
\usepackage{rotating, lscape}
\usepackage{setspace}
\usepackage{bbm}
\usepackage{natbib}
\usepackage{amsthm}
\usepackage{tikz}
\usepackage{etoolbox} 
\usepackage{listofitems} 
\usepackage{url}
\usepackage{appendix}
\usepackage{graphicx}
\usepackage{makecell}
\usepackage{bm}
\usepackage{bbm}
\usepackage{amsmath}
\usepackage{booktabs}
\usepackage{multirow}
\usepackage{arydshln}
\usepackage{enumitem}
\usepackage{eucal}
\usepackage{booktabs}
\usepackage{algorithm}
\usepackage{algpseudocode}
\usepackage{color}
\usepackage[colorlinks=true,
            linkcolor=darkgray,
            urlcolor=blue,
            citecolor=brown]{hyperref}
\graphicspath{ {./images/} }

\tikzset{>=latex} 
\colorlet{myred}{red!80!black}
\colorlet{myblue}{blue!80!black}
\colorlet{mygreen}{green!60!black}
\colorlet{mydarkred}{myred!40!black}
\colorlet{mydarkblue}{myblue!40!black}
\colorlet{mydarkgreen}{mygreen!40!black}
\tikzstyle{node}=[very thick,circle,draw=myblue,minimum size=22,inner sep=0.5,outer sep=0.6]
\tikzstyle{connect}=[->,thick,mydarkblue,shorten >=1]
\tikzset{ 
  node 1/.style={node,mydarkgreen,draw=mygreen,fill=mygreen!25},
  node 2/.style={node,mydarkblue,draw=myblue,fill=myblue!20},
  node 3/.style={node,mydarkred,draw=myred,fill=myred!20},
}

\theoremstyle{definition}

\newcommand{\bi}{\begin{itemize}}
	\newcommand{\ei}{\end{itemize}}
\newcommand{\bfs}{\mathbf{s}}
\newcommand{\pkg}[1]{\texttt{#1}}
\begin{document}
	\thispagestyle{empty} \baselineskip=28pt \vskip 5mm
	\begin{center} {\Large{\bf Spatio-temporal DeepKriging for Interpolation and Probabilistic Forecasting}}
	\end{center}
	
	\baselineskip=12pt \vskip 5mm
	
	\begin{center}\large
		Pratik Nag\footnote[1]{
			\baselineskip=10pt CEMSE Division, Statistics Program, King Abdullah University of Science and Technology, Thuwal 23955-6900, Saudi Arabia. E-mail: pratik.nag@kaust.edu.sa; ying.sun@kaust.edu.sa}, Ying Sun$^1$
			and Brian J Reich\footnote[2]{
			\baselineskip=10pt Department of Statistics, North Carolina State University, Raleigh, USA. Email: bjreich@ncsu.edu} 
    \end{center}
	
	\baselineskip=16pt \vskip 1mm \centerline{\today} \vskip 8mm
	
	\begin{center}
		{\large{\bf Abstract}} 
		\bi
	Gaussian processes (GP) and Kriging are widely used in traditional spatio-temporal mod-elling and prediction. These techniques typically presuppose that the data are observed from a stationary GP with parametric covariance structure. However, processes in real-world applications often exhibit non-Gaussianity and nonstationarity. Moreover, likelihood-based inference for GPs is computationally expensive and thus prohibitive for large datasets. In this paper we propose a deep neural network (DNN) based two-stage model for spatio-temporal interpolation and forecasting. Interpolation is performed in the first step, which utilizes a dependent DNN with the embedding layer constructed with spatio-temporal basis functions. For the second stage, we use Long-Short Term Memory (LSTM) and convolutional LSTM to forecast future observations at a given location. We adopt the quantile-based loss function in the DNN to provide probabilistic forecasting. Compared to Kriging, the proposed method does not require specifying covariance functions or making stationarity assumption, and is computationally efficient. Therefore, it is suitable for large-scale prediction of complex spatio-temporal processes. We apply our method to monthly $PM_{2.5}$ data at more than $200,000$ space-time locations from January 1999 to December 2022 for fast imputation of missing values and forecasts with uncertainties.
	\ei
	\end{center}
	\baselineskip=17pt

	\begin{doublespace}
		
		\par\vfill\noindent
	\par\medskip\noindent
		{\bf Short title}: Space-Time DeepKriging
	\end{doublespace}
	
	\clearpage\pagebreak\newpage \pagenumbering{arabic}
	\baselineskip=26.5pt
	

\section{Introduction}\label{chap:intro}

Statistical modeling of phenomena that evolve over space and time is critical in areas such as environmental monitoring and detecting climate change. 
Spatio-temporal data are often very large due to advances in data collection technology such as
sensors, satellites, model simulations and the internet of things. This presents significant challenges to conventional geostatistical approaches.
Interpolation and forecasting are key tasks in spatio-temporal data analysis. For instance, missing data caused by poor radar coverage due to atmospheric conditions or faulty sensors, unavailability of weather stations in particular areas, etc, must be imputed when employing environmental variables as inputs to climate and weather models \citep{zhang2018missing}. For forecasting, a probabilistic forecast takes the form of a predictive probability distribution over future quantities or events of interest. A comprehensive review on probabilistic forecasting can be found in \cite{gneiting2014probabilistic}. There exist numerous methods for spatio-temporal interpolation and forecasting in the literature, including numerical models, machine learning models, and geostatistical models \citep{fouilloy2018solar,bergen2019machine,zhang2019joint,irrgang2021will}, each having its advantages and disadvantages described below. 

In climate and weather applications, numerical models produce predictions via solving partial differential equations (PDEs) \citep{binkowski2003models,minet2020quantifying} which depict systems being studied in a physically-consistent manner. While the majority of spatial and spatio-temporal systems, however, display stochastic behavior, numerical approaches only produce deterministic solutions and do not account for uncertainty. 

Conventional statistical methods allow for both point prediction and associated uncertainties. In geostatistics, Kriging is the best linear unbiased predictor (BLUP) and is commonly used to provide point predictions as well as associated uncertainties. By assuming normality, the predictive distribution is fully specified here. Kriging for spatio-temporal data requires a covariance model. A comprehensive review of space-time covariance structures and models can be found in \cite{montero2015spatial} and \cite{chen2021space}.  A major limitation of Kriging, however, is the computational cost, especially for spatio-temporal datasets \citep{sun2012geostatistics}.  
Along with covariance models, a variety of modeling approaches have been proposed for spatio-temporal processes in the last two decades for stationary processes \citep[e.g.,][]{stein2005space,bartlett2013statistical}. Several frameworks have also been presented for prediction of nonstationary spatio-temporal processes, such as, 
\cite{wikle1998hierarchical,stroud2001dynamic,ma2002spatio,huang2004modeling,kolovos2004methods,fuentes2008class,bruno2009simple,sigrist2012dynamic,xu2018improved}. \cite{cressie2015statistics} gives a comprehensive overview on spatio-temporal modelling for stationary and nonstationary data through classical statistical approaches as well as recently developed approaches such as hierarchical dynamical spatio-temporal models. Computational load for expanding most of these methods to large datasets, however, is very high.

 In recent years, deep neural networks (DNN) are being widely used in environmental applications. In addition to being nonparametric, DNNs are computationally efficient and can be applied to large datasets~\citep{najafabadi2015deep}. Several researchers have incorporated DNNs in spatio-temporal forecasting. Recurrent neural networks (RNN), specifically Long-Short Term Memory Networks (LSTM), have been applied to various forecasting tasks \citep{elsworth2020time,cao2019financial}. An alternative representation of RNN with the basis of reservoir computing \citep{lukovsevivcius2009reservoir} is the echo state network (ESN) \citep{jaeger2007echo}. \cite{mcdermott2017ensemble} used ESNs to produce long lead forecasts and quantified the uncertainties with ensamble approach for complex spatio-temporal processes. More recently, \cite{https://doi.org/10.48550/arxiv.2102.01141} used a spatio-temporal ESN to forecast short-term wind speeds for power production over Saudi Arabia with high-resolution wind fields. Moreover \cite{graph_NN_st} used graph neural networks for space-time forecasting of the wind fields. \cite{liu2014deep} proposed a new approach for feature representation for weather forecasting. Several other methods, such as precipitation modelling \citep{kuhnlein2014improving}, detecting extreme weather events \citep{liu2016application} and wind speed forecasting \citep{liu2018smart}, have been developed for spatial interpolation and forecasting. 
 
 Convolutional neural networks (CNN) can effectively capture spatial dependency for regularly gridded data. \cite{shi2015convolutional} used the Convolutional-LSTM for spatio-temporal forecasting. \cite{zammit2020deep} used the CNNs in a hierarchical statistical Integro-Difference Equation (IDE) framework for probabilistic forecasting. 
 
 For irregularly gridded spatial data, \cite{chen2020deepkriging} used the basis functions to capture the spatial dependence and use them as features in the DNN. \cite{amato2020novel} proposed a fresh strategy for using DNNs to spatio-temporal modeling. They employed basis functions as inputs for spatio-temporal interpolation after decomposing the spatio-temporal processes into a sum of products of temporally referenced basis functions. A comprehensive review on more recent statistical and deep learning frameworks for spatio-temporal forecasting can be found in \cite{https://doi.org/10.48550/arxiv.2206.02218}. Nevertheless, the majority of these algorithms either do not provide prediction uncertainty or may not be appropriate for irregularly gridded spatial data. By offering a thorough modeling framework for spatio-temporal interpolation and probabilistic forecasting, our proposed approach attempts to address these issues.

This paper is an extension of \cite{chen2020deepkriging} for spatio-temporal modelling. We implement a two-stage framework for interpolation and forecasting. In the first stage, we use spatio-temporal basis functions to embed the coordinates for spatio-temporal interpolation. We employ the Gaussian kernel to build the temporal bases, and radial basis functions \citep{buhmann2000radial} to capture spatial trends. In the second stage, we use convolutional LSTM to forecast future observations. We incorporate a quantile-based loss function for probabilistic forecasting. The proposed method is non-parametric, making it more adaptable and appropriate for non-Gaussian and non-stationary spatio-temporal processes. Unlike likelihood-based inference, it does not require Cholesky factorization of large covariance matrices and thus the approach is computationally inexpensive. 


\section{Methodology}\label{sec:method}

\subsection{Overview}

Consider the real valued spatio-temporal random field \{$Y(\mathbf{s},t), \mathbf{s}\in D, t \in \mathcal{T}\}, D \subseteq \mathbb{R}^{p}, \mathcal{T} \subseteq \mathbb{R} $, with realizations $ \mathbf{Z}_{N,K} =  \{Z(\mathbf{s}_1,t_1),{Z}(\mathbf{s}_2,t_1),...,{Z}(\mathbf{s}_N,t_K)\}$ at $N$ spatial locations and $K$ time points. Assuming a classical additive model one can write 

\begin{equation}
        Z(\mathbf{s},t) = Y(\mathbf{s},t) + \epsilon(\mathbf{s},t),
        \label{eq:obs}
    \end{equation}
where ${\epsilon(\mathbf{s},t)}$, known as nugget effect, is the univariate white noise with variance $\sigma^2(\mathbf{s},t)$ independent over $\bfs$ and $t$. For Gaussian Random fields \citep{bardeen1986statistics}, Kriging \cite[Ch.\ 6, p.\ 321]{cressie2015statistics} is most often used for spatio-temporal prediction. Here  
${Y}(\mathbf{s},t)$ can be modelled as
\begin{equation}
    {Y}(\mathbf{s},t) = \mu (\mathbf{s},t) +{\gamma(s,t)},
    \label{eq:grf}
\end{equation}
where $\mu(\mathbf{s},t)$ is the mean process usually assumed to be a linear combination of the known covariates $X_{vec}(\bfs,t)$ and ${\gamma}(\mathbf{s},t)$ is a zero-mean spatio-temporal random process with cross-covariance function $\mathcal{C}(\gamma(s,t),\gamma(s',t'))$. Several classes of stationary covariance functions have been developed \citep{cressie1999classes} for modelling space-time interactions, such as the Mat\'ern space-time covariance function \citep{gneiting2002nonseparable}, stationary fully symmetric covariance models based on spectral methods and process mixtures~\citep{de2001space,gneiting2002nonseparable,porcu2008new} etc. Estimation of the covariance parameters of these functions usually relies on the maximum likelihood estimation (MLE) approach. The computation of this MLE, however, requires $\mathcal{O}(K^2N^2)$ memory and $\mathcal{O}(K^3N^3)$ operations per iteration due to the Cholesky decomposition of the $KN \times KN$ covariance matrix which for large $N$ and $K$ can be computationally expensive. \cite{9397281} encountered this problem with the help of high performance parallel computation (HPC), however, their implementation is only limited to stationary Gaussian processes. \cite{blangiardo2013spatial} reviewed the R-package R-INLA for spatial and spatio-temporal modelling through Integrated Nested Laplace Approximation (INLA) approach and Stochastic Partial Differential Equation (SPDE) approach. \cite{zammit2021frk} developed an R package \pkg{FRK} based on fixed rank kriging \citep{cressie2008fixed} for spatial/spatio-temporal kriging for large datasets. In this paper, we propose to use multi-layer deep neural networks (DNNs) to create flexible and nonlinear models for space-time interpolation and probabilistic forecasting, which is suitable for large-scale prediction of nonstationary and non-Gaussian processes.

In the remainder of this section, we address two common objectives associated with spatio-temporal modelling namely interpolation and forecasting. In Section \ref{sec:model1} we briefly describe the problems of interpolation and forecasting and define spatio-temporal basis function and a feed forward NN for interpolation. This is extended in Section \ref{sec:model2} using a LSTM network for forecasting. 

\subsection{Spatio-temporal interpolation}\label{sec:model1}

Given observations $\mathbf{Z}_{N,K}$, two common goals of spatio-temporal prediction are interpolation, i.e., predict the true process $Y(\mathbf{s}_0,t)$ at unobserved spatial location $\mathbf{s}_0$, and forecasting, i.e., predict  $Y(\bfs,t_{K+u})$ at unobserved location $\bfs_0$ or observed location $\mathbf{s}$ at a future time point $t_{K+u}$. 

For the first scenario, the optimal predictor with parameter set $\boldsymbol{\theta}$ can be defined by minimizing the expected value of the loss function, that is, 
 \begin{equation}
       \hat{Y}^{opt}(\mathbf{s}_0,Z) = \mathbb{E}\{L(\hat{{Y}}(\mathbf{s}_0,t),{Y(s_0,t)})| \mathbf{Z}_{N,K}\} = \operatorname*{argmin}_{{\hat{Y}}} R_1(\hat{{Y}}(\mathbf{s_0},t | \mathbf{Z}_{N,K}),{Y(\mathbf{s}_0,t)}),
       \label{eq:1}
    \end{equation}
where $\hat{{Y}}^{opt}(\mathbf{s_0},{Z})$ is the optimal predictor given $\mathbf{Z}_{N,K}$. The function $R_1(.,.)$ is the risk function which is the expected value of the loss $L$. Note that \eqref{eq:1} is for spatio-temporal interpolation and not forecasting. 

For forecasting, the optimal predictor can be defined as 
 \begin{equation}
       \resizebox{.9\hsize}{!}{$\hat{Y}^{opt}(\mathbf{s}_0,t_{K+u},Z) = \mathbb{E}\{L(\hat{{Y}}(\mathbf{s}_0,t_{K+u}),{Y(s_0,t_{K+u})})| \mathbf{Z}_{N,K}\} = \operatorname*{argmin}_{{\hat{Y}}} R_2(\hat{{Y}}(\mathbf{s_0},t_{K+u}| \mathbf{Z}_{N,K}),{Y(\mathbf{s}_0,t_{K+u})})$},
       \label{eq:2}
    \end{equation}
 where $\hat{Y}^{opt}(\mathbf{s}_0,t_{K+u},Z)$ is the optimal predictor with risk function $R_2(.,.)$.

 DeepKriging, as proposed by \cite{chen2020deepkriging}, tackles these problems for univariate spatial processes by formulating the problem as regression and introducing a spatially dependent neural network structure for spatial prediction. To enable the DNNs to accurately represent the geographical dependence of the process, they used the radial basis functions to incorporate the spatial positions into a vector of weights.
Similar to \cite{chen2020deepkriging}, we suggest using spatio-temporal basis functions for the space-time processes to generate the embedding layer as opposed to directly feeding DNN the coordinates of the observations. This idea is motivated by Karhunen-Lo\'eve expansion \citep{daw2022overview} of univariate random field i.e., a spatio-temporal process ${\gamma(\mathbf{s},t)}$ can be represented as  
       ${\gamma(\mathbf{s},t)} = \sum_{p=1}^{\infty} w_{p}\boldsymbol{\phi}_{p}(\mathbf{s},t)$,
    where the $w_{p}$'s are independent random variables and $\boldsymbol{\phi}_{p}(\mathbf{s},t)$ are the pairwise orthonormal spatio-temporal basis functions. 
    Therefore, a spatio-temporal process can be approximated by the finite sum
    $
    {\gamma(\bfs,t)} \approx \sum_{p=1}^{Q} w_{p}\boldsymbol{\phi}_{p}(\mathbf{s},t),
    $
    where the weights $w_{p}$ can be estimated by minimizing the risk function $R_1$. Here $\boldsymbol{\phi}_{p}(\mathbf{s},t)$ can be obtained by the tensor-product of the spatial bases $\phi_i(\bfs), i = 1,2,...,G$ and temporal bases $\psi_j(t), j=1,2,...,H $ \citep{cressie2008fixed} and thus includes $G H$ basis functions. Therefore, the interpolation problem can be viewed as linear regression \citep{borchani2015survey} with these basis functions $\boldsymbol{\phi}_{p}(\mathbf{s},t)$ as covariates.

We have selected the multi-resolution compactly supported Wendland radial basis function  \citep{nychka2015multiresolution}  from a large number of existing basis functions including spline basis functions \citep{wahba1990spline}, wavelet basis functions \citep{vidakovic2009statistical}, and radial basis functions \citep{hastie2001elements} for spatial embedding. The Wendland basis functions are defined via $B_1(d) = \frac{(1-d)^6}{3}(35d^2 + 18d + 3)\mathbf{1}\{0\leq d \leq 1 \}$. The spatial basis functions are then defined as $\phi_i(\bfs) = B_1(\left\|\bfs-u_i\right\|/\theta)$ with $\theta$ as the bandwidth parameter and anchor points (spatial locations) $\{u_1,u_2,...,u_{G} \}$. We take $\{u_1,u_2,...,u_{G} \}$ to be a square grid of locations covering the spatial domain and $\theta$ to be 2.5 times the distance between adjacent anchor points. To extract the temporal dependence we have used the Gaussian radial basis functions over the time domain given as $\psi_j(t) = \exp(-0.5 (t-v_j)^2/(\kappa^2)) $ with anchor points (time points) $v \in \{v_1,v_2,...,v_H \}$ and scale set to $\kappa=| v_1 - v_2|$. One can create numerous sets of basis functions by choosing various $G$ and $H$ values. We will be able to explain both the long and short-range spatial and temporal dependencies as a result.
    
These coordinate embeddings can now be used as inputs to the DNN to capture the spatial aspects of the data. To reduce the computational burden, instead of calculating the tensor-product, we have stacked the spatial and temporal basis functions together to get $\boldsymbol{\phi}(\mathbf{s},t) = \{ \phi_1(\bfs),...,\phi_G(\bfs), \psi_1(t),..., \psi_H(t)\}^T$, where the cardinality of $\boldsymbol{\phi}(\mathbf{s},t)$ is $Q = G+H$. Note that this choice of the basis functions do not imply separability in space and time because the weights are shared by each node in deep neural networks. That is, deep learning is capable of learning interactions between spatial and temporal basis functions.
    
\begin{figure}[ht]
\centering
\includegraphics[scale=0.5]{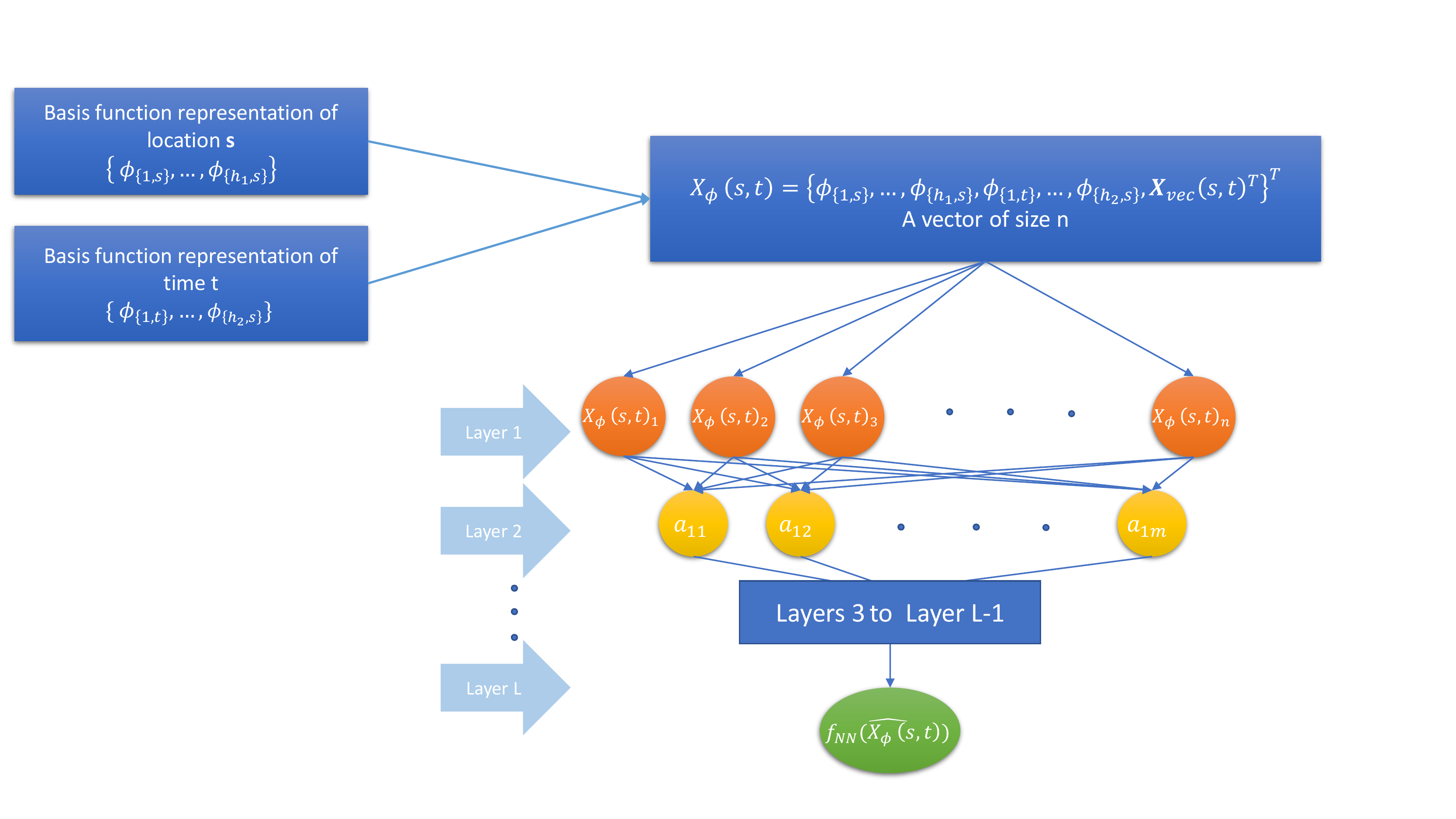}
\caption{Architecture of the feed-forward neural network for spatio-temporal interpolation}
\label{fig:NN_structure}
\end{figure}

We have used a single-output deep neural network structure to build the spatio-temporal DeepKriging framework. We define $\mathbf{X}_{\phi}(\mathbf{s},t)$ as the vector of inputs containing the embedded vector of basis functions $\boldsymbol{\phi}(\mathbf{s},t)$ and the covariates $\mathbf{X}_{vec}(\mathbf{s},t)$, Then $\mathbf{X}_{\phi}(\mathbf{s},t) = (\boldsymbol{\phi}(\mathbf{s},t)^T,\mathbf{X}_{vec}(\mathbf{s},t)^T)^T$. Taking $\mathbf{X}_{\phi}(\mathbf{s},t)$ as the inputs to a neural network with L layers and $M_l$ nodes in layer $l=1,...,L$ DNN can be specified as, 
        
    \begin{equation}
    \begin{aligned}
        h_1(\mathbf{X}_{\phi}(\mathbf{s},t)) &= W_1\mathbf{X}_{\phi}(\mathbf{s},t) + b_1, \ a_1(\mathbf{s},t)=\psi(h_1(\mathbf{X}_{\phi}(\mathbf{s},t))) \\
        h_2(a_1(\mathbf{s},t)) &= W_2a_1(\mathbf{s},t) + b_2, \ a_2(\mathbf{s})=\psi(h_2(a_1(\mathbf{s},t))) \\
        . . &. \\
        h_L(a_{L-1}(\mathbf{s},t)) &= W_La_{L-1}(\mathbf{s},t) + b_L, \ f_{NN_{\tau}}(\mathbf{X}_{\phi}(\mathbf{s},t))=\Psi(\tau,h_L(a_{L-1}(\mathbf{s},t))), \\
    \label{eq:6}
    \end{aligned}
    \end{equation}
where $f_{NN_{\tau}}(\mathbf{X}_{\phi}(\mathbf{s},t))$ is the prediction output from the model, the weight matrix $W_l$ is a $M_l \times M_{l-1}$ matrix of parameters and $b_l$, a $M_l \times 1$ vector is the bias at layer $l$. The parameter set of this network is  $\boldsymbol{\theta}_1 = \{ W_l, b_l : l = 1,2,...,L\}$. In \eqref{eq:6}, $\psi(.)$ signifies the activation function which in our case is the rectified linear unit ReLU \citep{schmidt2020nonparametric}. We will discuss more about the function $\Psi(\tau,.)$ in the following paragraphs. The spatio-temporally dependent neural network, or \textbf{Space-Time.DeepKriging}, topology is graphically depicted in Figure  \ref{fig:NN_structure}. 

The quantile regression of \cite{koenker1978regression} is popularly used as an alternative to conventional regression to learn the relationship between response and predictor variables. By simultaneously estimating multiple quantiles, quantile regression makes it possible to produce relevant summaries of the conditional distribution of the response variable for specific predictor factors \citep{koenker2004quantile,li20081}. \cite{moon2021learning} extended this framework for neural networks to learn quantiles through nonlinear regression. However, their approach is restricted to shallow neural network architecture only. In this paper we have extended this work to provide quantiles for spatio-temporal forecasts.

\cite{koenker1994quantile} proposed the linear model for $\tau$-quantile function for a given $\tau \in (0,1)$. In this paper we use the nonlinear function $f_{NN_{\tau}}(\mathbf{X}_{\phi}(\mathbf{s},t))$ for quantile prediction. Here we implement stochastic gradient descent \citep{amari1993backpropagation,ketkar2017stochastic} for minimization of the empirical version of the risk function $R_1(.,.)$ based on $\boldsymbol{\theta}_1$ given as

\begin{equation}
    R_1\{f_{NN_{\tau}}(\mathbf{X}_{\phi}(\mathbf{s},t);\boldsymbol{\theta}_1),\mathbf{Z}_{N,K}\} = \frac{1}{K}\sum_{k=1}^K \rho_{\tau}
    (f_{NN_{\tau}}(\mathbf{X}_{\phi}(\mathbf{s}_n,t_k))-Z(\mathbf{s}_n,t_k)),
    \label{eq:5}
\end{equation}
where $\rho_{\tau}(v) = v(\tau - I(v<0))$ is the check loss function.

By the definition of quantiles, the conditional quantile functions should not be crossing. That is, $f_{NN_{\tau_1}}(\mathbf{X}_{\phi}(\mathbf{s},t)) \leq f_{NN_{\tau_2}}(\mathbf{X}_{\phi}(\mathbf{s},t))$ should be satisfied for all $\bfs_0$ and $t$ when $\tau_1 \leq \tau_2$. To assure this we have constrained the output of $f_{NN_{\tau}}(\mathbf{X}_{\phi}(\mathbf{s},t))$ by the following choices of $\Psi$ for different values of $\tau$,
 
 \begin{equation}
  \Psi(\tau,x) =
    \begin{cases}
      x & \text{for $\tau = 0.5$} \\
      f_{Constant} + \frac{\lambda (\tau-0.5)}{1+e^{-x}} & \text{for $\tau > 0.5$} \\
      f_{Constant} - \frac{\lambda (0.5 - \tau)}{1+e^{-x}} & \text{for $\tau < 0.5 $} ,\\
    \end{cases}  
    \label{eq:psi}
\end{equation}
 where $f_{Constant} = \widehat{f_{NN_{0.5}}}(\mathbf{X}_{\phi}(\mathbf{s},t))$ is the \textbf{Space-Time.DeepKriging} model estimate for quantile level $0.5$. The hyper-paramter $\lambda$ controls the deviation of upper or lower quantiles from its median. One can choose $\lambda \propto \sigma_{range}/2$ where $\sigma_{range} = \text{max }{\mathbf{Z}_{N,K}} - \text{min }{\mathbf{Z}_{N,K}}$. By the architecture the output of $\Psi_{\tau}$ will always give us non-crossing quantile prediction intervals as the upper limit of the interval will always be greater than that of the median, whereas the lower limit will always be smaller. To obtain an $100(1-\alpha)$\% prediction interval we first need to get the estimate of median $f_{NN_{0.5}}(\mathbf{X}_{\phi}(\mathbf{s},t))$ and use that to compute $f_{NN_{1-\alpha/2}}(\mathbf{X}_{\phi}(\mathbf{s},t))$ and $f_{NN_{\alpha/2}}(\mathbf{X}_{\phi}(\mathbf{s},t))$ sequentially with $\Psi(\tau,x)$ as activation for the output layer. Hence the final predictor of the neural network at an unobserved location $\mathbf{s}_0$ is $\widehat{f_{NN_{\tau}}}(\mathbf{X}_{\phi}(\mathbf{s_0},t))$ where 
    \\
    $$
    \hat{\boldsymbol{\theta}}_1 = \operatorname*{argmin}_{\boldsymbol{\theta}_1} R_1\{f_{NN_{\tau}}(\mathbf{X}_{\phi}(\mathbf{s},t);\boldsymbol{\theta}_1),\mathbf{Z}_{N,K}\}.
    $$

\subsection{Probabilistic forecasting}\label{sec:model2}

Forecasting is one of the most crucial components of time series models. However, as mentioned in Section \ref{sec:model1}, \textbf{Space-Time.DeepKriging} is designed for interpolation, not forecasts at future time points. Therefore, we developed a second modeling framework to predict at future time points at unobserved locations that is conditioned on \textbf{Space-Time.DeepKriging}. 

\begin{figure}[ht]
\centering
\includegraphics[scale=0.5]{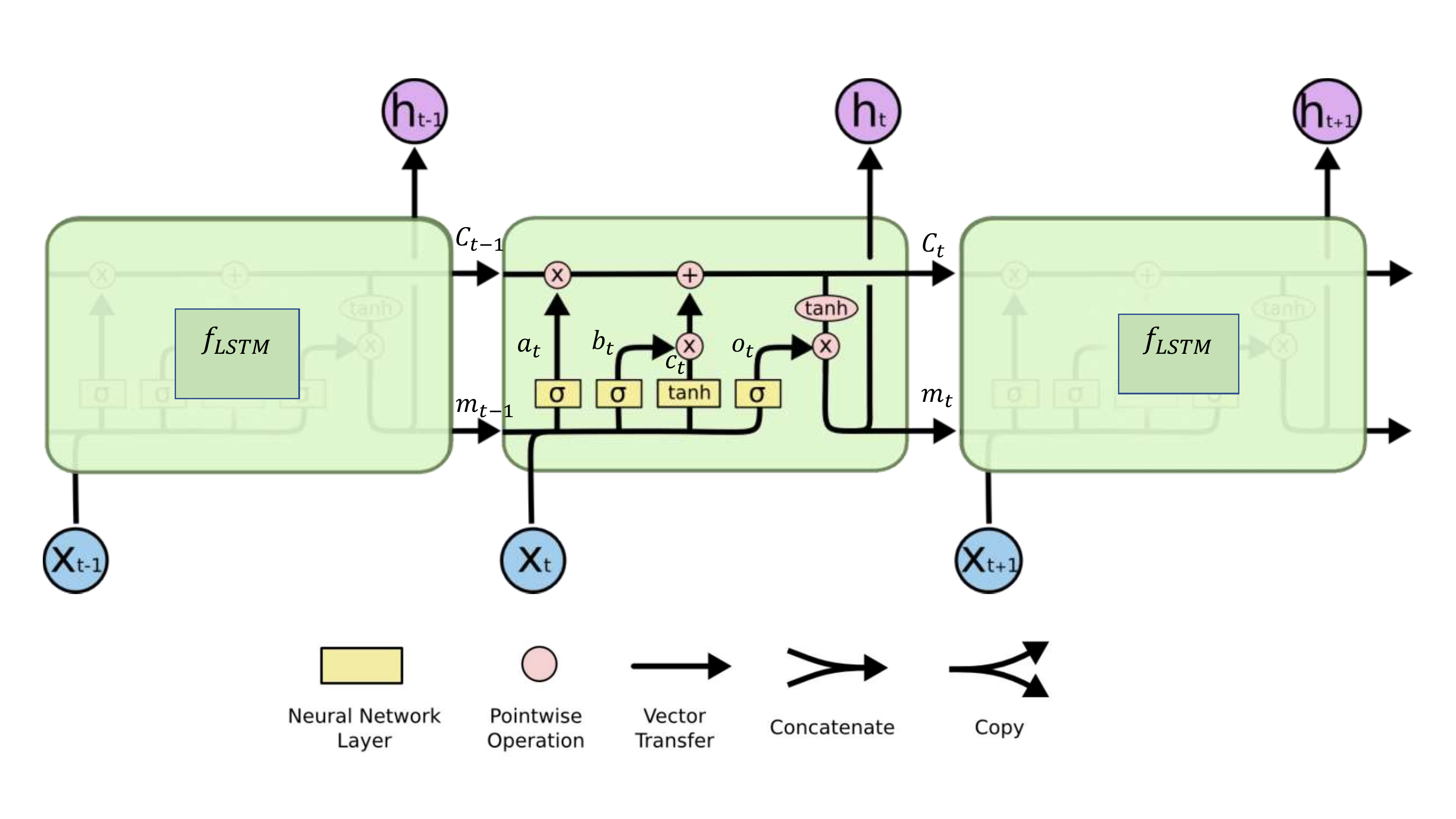}
\caption{Architecture of LSTM network used for forecasting}
\label{fig:LSTM}
\end{figure}

Long-short term memory network (LSTM) is a popular approach for time series forecasting in deep learning  \citep{cao2019financial,sagheer2019time,chimmula2020time}. LSTMs introduced by \cite{hochreiter1997long} are a special kind of Recurrent Neural Network (RNN) \citep{medsker2001recurrent} capable of learning long-term dependencies. Figure \ref{fig:LSTM} shows a general structure of the LSTM method. We first describe the single layer of a LSTM model for an arbitrary (non-spatial) time series problem with $X_t$ as the features observed at time $t$, and $Y_t$ as the response. $Y_t = f_{LSTM}(X_t) + e_t$, where $f_{LSTM}$ is the prediction and $e_t$ is error. The network is defined as 
$$
\begin{aligned}
    a_t = \psi_{\sigma}(W_a.[m_{t-1},X_t]) + b_a, \ &\ \ 
    b_t = \psi_{\sigma}(W_b.[m_{t-1},X_t]) + b_b, \\
    c_t = \psi_{tanh}(W_c.[m_{t-1},X_t]) + b_c, \ &\ \ 
    C_t = a_t \odot C_{t-1} + b_t \odot c_t, \\
    o_t = \psi_{\sigma}(W_o.[m_{t-1},X_t]) + b_o, \ &\  \ 
    m_t = o_t \odot \psi_{tanh}(C_t),
\end{aligned}
$$
where $\psi_{\sigma}(.)$ \citep{pratiwi2020sigmoid} and $\psi_{tanh}(.)$ \citep{lau2018review} are the sigmoid and tanh activation functions respectively. `$ \odot $' defines the Hadamard product between two vectors. Here $a_t, b_t, c_t$ and $o_t$ are four single layer neural networks with specific nonlinear activations. The other functions are multiplicative operations on these 4 layers. In a particular LSTM layer we take three inputs, i.e., two of which are the information from the last layer $C_{t-1},m_{t-1}$ and covariates for current layer $X_t$. $C_t$ is called the cell state which carries information from past states with little intervention by the current step. For more details on the functionalities of each of these functions see \cite{hochreiter1997long}. Hence
the output of a specific LSTM layer is written as $f_{LSTM}(X_t) = m_t$. For a $P$ layer LSTM stack, i,e., $P$ layers of the LSTM framework, where the output of layer $(i-1)$ is the input of layer $i$, $\theta_{LSTM_i} = \{W_{a}^i,W_{b}^i,W_{c}^i,W_{o}^i,b_{a}^i,b_{b}^i,b_c^i,b_o^i\}$ is the matrix of weights at the $i$-th LSTM layer. 

Returning to the spatio-temporal setting, we define the time series $\mathbf{X}^{NN}$ as the vector of inputs which contains the predictions from \textbf{Space-Time.DeepKriging} for location $\mathbf{s}_0$ at all observed time points i.e; $\mathbf{X}^{NN} = \{ \widehat{f_{NN_{\tau}}}(\mathbf{X}_{\phi}(\mathbf{s_0},t_1)), \cdots, \widehat{f_{NN_{\tau}}}(\mathbf{X}_{\phi}(\mathbf{s_0},t_K))\}$. For a given $\tau \in (0,1)$ we use this LSTM-based framework for predicting the quantile-based forecasts (\textbf{QLSTM}). With $P$ hidden layers the network can be specified as 

    \begin{equation}
        \begin{aligned}
            h_1(\mathbf{s}_0,t) &= f_{LSTM}(\{\mathbf{X}^{NN}_{t-j-1}, \cdots, \mathbf{X}^{NN}_{t-1}\}) , \ \text{$\mathbf{X}^{NN}_{j}$ is the j-th element in $\mathbf{X}^{NN}$,} \\
            h_2(\mathbf{s}_0,t) &= f_{LSTM}(h_1(\mathbf{s}_0,t))\\
            . . &. \\
            h_{P-1}(\mathbf{s}_0,t) &= f_{LSTM}(h_{P-2}(\mathbf{s}_0,t)) \\
            h_P(\mathbf{s}_0,t) &= W_P h_{P-1}(\mathbf{s}_0,t) + b_P, \ f_{NN_{\tau}}^{LSTM}(\mathbf{s}_0,t) = \Psi(\tau,h_P(\mathbf{s}_0,t)),\\
        \end{aligned}
    \label{eq:forecast}
    \end{equation}
where $f_{NN_{\tau}}^{LSTM}(\mathbf{s}_0,t)$ is the prediction output from model at quantile level $\tau$, the last layer is a fully connected feed forward neural network layer with weight matrix $W_P$ with dimension $M_P \times M_{P-1}$ and $b_P$ is the bias, and $\boldsymbol{\theta}_{LSTM} = \{\theta_{LSTM_1},...,\theta_{LSTM_{P-1}},W_P,b_P\}$ is the set of trainable parameters. $\Psi(\tau,.)$ is the function defined in \eqref{eq:psi} with $f_{Constant} = f_{NN_{0.5}}^{LSTM}(\mathbf{s}_0,t)$.

Implementing similar architecture for the loss function as discussed in Section \ref{sec:model1}, the quantile loss function associated with \textbf{QLSTM} can be written as 

\begin{equation}
    L_{\tau}\{f_{NN_{\tau}}^{LSTM}((\mathbf{s}_0,t);\boldsymbol{\theta}_{LSTM}),\mathbf{X}^{NN}\} = \frac{1}{K}\sum_{k=1}^K \rho_{\tau}
    (f_{NN_{\tau}}^{LSTM}(\mathbf{s}_0,t_k)-\mathbf{X}^{NN}_k).
    \label{eq:loss_LSTM}
\end{equation}

Note that, an empirical version of the risk function defined in \eqref{eq:2} can be written as \\
$R_2\{f_{NN_{\tau}}^{LSTM}((\mathbf{s}_0,t);\boldsymbol{\theta}_{LSTM}),\mathbf{Z}_{\mathbf{s}_0,K}\}$ where $ \mathbf{Z}_{\mathbf{s}_0,K} = \{{Z(\mathbf{s}_0,t_1)},{Z(\mathbf{s}_0,t_2)}, \cdots ,{Z(\mathbf{s}_0,t_K)}\}$. However, we do not have the vector $\mathbf{Z}_{\mathbf{s}_0,K}$ as location $\mathbf{s}_0$, that is unobserved, hence we estimate $\mathbf{Z}_{\mathbf{s}_0,K}$  with $\mathbf{X}^{NN}$, the interpolated time series. So the risk function $R_2(.,.)$ can be thus approximated by $L_{\tau}(.,.)$. 
Algorithm \ref{algo:1} provides the pseudocode for the approach.
 
 \textbf{QLSTM} is a distribution-free quantile forecast model. However, this strategy has a significant flaw in that as it fails to take spatial dependence into account when making forecasts. To overcome, we present the Convolutional Long-Short Term Memory Network \citep{shi2015convolutional}, which instead of feed forward networks takes the 
 Convolution operation output as input to the LSTM. A convolutional layer can be defined as follows \citep{o2015introduction}:
This layer extracts the spatial neighborhood structure from the data. The input is observed on $r\times r$ regular grid. For our case the kernel size for the convolution layer is $3\times 3$. The output is set to have 64 feature maps, where each feature contains $(r-2)\times (r-2)$ values. 
 
 Let $\mathbf{X}^{NN_{CONV}} = \{ \mathcal{A}(\mathbf{s}_0,t_1), \ldots ,\mathcal{A}(\mathbf{s}_0,t_K)\}$. The matrix $ \mathcal{A}(\mathbf{s}_0,t) = \{ \widehat{f_{NN_{\tau}}}(\mathbf{X}_{\phi}(\mathbf{s}_j,t) ;\hat{\boldsymbol{\theta}_1}) : \mathbf{s}_j \in N_{\mathbf{s}_0}\}$, where $N_{\mathbf{s}_0}$ is the neighbourhood of $\mathbf{s}_0$, is a $r \times r$ matrix with elements $[X_t(i, j)]_{i, j \in \{1, \ldots, r\}}$. Then the output of one convolution layer is $[m_{i, j}^{(\eta)}]_{i, j \in \{2, \ldots, r-2\}}$, where $\eta \in \{1, \ldots, 64\}$ for each feature map. $m_{i, j}^{(\eta)}$ is determined by 
	\begin{equation*}
		m_{i, j}^{(\eta)} = \left\{ \sum_{k_1 = i-1}^{i+1} \sum_{k_2 = j-1}^{j+1} X_t(k_1, k_2) \mu({k_1-i,k_2-j})^{(\eta)} \right\},
	\end{equation*}
	where $\mu({k_1-i,k_2-j})^{(\eta)} $ are elements of the $\eta$-th filter which is a $3 \times 3$ matrix and $\theta_{conv} = \{\mu({k_1-i,k_2-j})^{(\eta)}, i, j \in \{2, \ldots, r-2\}, \eta \in \{1, \ldots, 64\}\}$ are parameters in this layer. 
	
	Using the convolution as inputs to the LSTM we can rewrite \eqref{eq:forecast} as 
 
\begin{equation}
        \begin{aligned}
            h_1(\mathbf{s}_0,t) &= f_{ConvLSTM}(\{\mathbf{X}^{NN_{Conv}}_{t-j-1}, \cdots, \mathbf{X}^{NN_{Conv}}_{t-1}\}) , \ \text{$\mathbf{X}^{NN_{Conv}}_{j}$ is the j-th element in $\mathbf{X}^{NN_{Conv}}$,} \\
            h_2(\mathbf{s}_0,t) &= f_{ConvLSTM}(h_1(\mathbf{s}_0,t))\\
            . . &. \\
            h_{P-1}(\mathbf{s}_0,t) &= f_{ConvLSTM}(h_{P-2}(\mathbf{s}_0,t)) \\
            h_P(\mathbf{s}_0,t) &= W_P h_{P-1}(\mathbf{s}_0,t) + b_P, \ f_{NN_{\tau}}^{Conv}(\mathbf{s}_0,t) = \Psi_{\tau}(h_P(\mathbf{s}_0,t)),\\
        \end{aligned}
    \label{eq:forecast_conv}
    \end{equation}
where $f_{ConvLSTM}(.)$ is the Convolutional LSTM layer with parameter vector $\boldsymbol{\theta}_{ConvLSTM}$. We call this framework as \textbf{QConvLSTM}.  

Hence the final predictor with \textbf{QLSTM} at a future time point $t_{K+u}$ is $\widehat{f_{NN_{\tau}}^{LSTM}}((\bfs_0,t_{K+u});\hat{\boldsymbol{\theta}}_{LSTM})$ where 
    \\
    $$
    \hat{\boldsymbol{\theta}}_{LSTM} = \operatorname*{argmin}_{\boldsymbol{\theta}_{LSTM}} L_{\tau}\{f_{NN_{\tau}}^{LSTM}((\mathbf{s}_0,t);\boldsymbol{\theta}_{LSTM}),\mathbf{X}^{NN}\}.
    $$
For \textbf{QConvLSTM}, the predictor is $\widehat{f_{NN_{\tau}}^{Conv}}((\bfs_0,t_{K+u});\hat{\boldsymbol{\theta}}_{ConvLSTM})$ where 
    \\
    $$
    \hat{\boldsymbol{\theta}}_{ConvLSTM} = \operatorname*{argmin}_{\boldsymbol{\theta}_{ConvLSTM}} L_{\tau}\{f_{NN_{\tau}}^{Conv}((\mathbf{s}_0,t);\boldsymbol{\theta}_{ConvLSTM}),\mathbf{X}^{NN_{Conv}}\}.
    $$

\begin{algorithm}[ht]
\caption{Algorithm \textbf{QLSTM}}\label{alg:cap}
\begin{algorithmic}
\State input : Interpolation vector $\mathbf{X}^{NN}$.
\State output : forecast for future time points at location $\mathbf{s}_0$.

\For{k in $1,2,...,K$}
    \State input vector : $\{\mathbf{X}^{NN}_{k-j-1}, \cdots ,\mathbf{X}^{NN}_{k-1} \}$ (length $j$)
    \State input response : $\mathbf{X}^{NN}_k$
    \State output : $f_{NN_{\tau}}(\mathbf{s}_0,k)$
    \State Minimize $L_{\tau}\{f_{NN_{\tau}}((\mathbf{s}_0,t);\boldsymbol{\theta}_{LSTM}),\mathbf{X}^{NN}\}$ through backpropagation.
\EndFor
\State{Predict at time point $K+1$ using $\{\mathbf{X}^{NN}_{K-j},...,\mathbf{X}^{NN}_K\}$ }
\end{algorithmic}
\label{algo:1}
\end{algorithm}

Algorithm for \textbf{QConvLSTM} is also similar. Here we use input vector $\mathbf{X}^{NN_{Conv}}$ which gives output $f_{NN_{\tau}}^{Conv}(\mathbf{s}_0,t)$. 
\section{Simulation studies}\label{sec:prediction2}

We have made two separate simulation studies. The details of each simulation with corresponding results have been provided in the subsections below. 
\subsection{KAUST competition data}\label{sec:competition}
We have used our proposed approach in the KAUST spatial statistics competition for spatio-temporal prediction \citep{nag_2022}. The datasets were simulated using \textbf{ExaGeoStat} \citep{9397281} from a zero-mean Gaussian process with a non-separable stationary space-time covariance function at space-time locations $(\bm s,t)\in [0,1]^{2}\times \mathbb{R}$ :
\begin{equation}
    C(\bm h,v)= \frac{\sigma^2}{a_{t}|v|^{2\alpha}+1}
    {\cal{M_\mathcal{\nu}}} \left\{\frac{\|\bm h\|/a_{s} }{(a_{t}|v|^{2\alpha}+1)^{\beta/2} }\right\}, \label{eq:st-cov}
\end{equation}
where $\sigma^2> 0$ is the variance, $\nu >0 $ and $\alpha \in (0,1]$ are the smoothing parameters, $a_{s}$ and $a_{t}>0$ are the range parameters in space and time, respectively, $\beta \in (0,1]$ is the space-time interaction parameter, $\bm h$ and $v$ are the distances between two spatial points and two time points respectively, and ${\cal{M_\mathcal{\nu}}}$ is the univariate Mat\'ern correlation function. Eighteen separate univariate space-time datasets were generated with different parameter settings, out of which nine were of size 90K (sub-competition 2a) and the rest were of size 1M (sub-competition 2b). Details on the parameter settings can be found in Table 1 in \cite{nag_2022}.
In sub-compatition 2a, each dataset had 100 time stamps and 900 locations per time stamp. For sub-compatition 2b, each dataset had 100 time stamps and 10,000 locations per time stamp. Three different scenarios were considered for leaving out space-time points for prediction:

 \begin{itemize}
        \item[1.] Random spatial locations with all times left out; 
        
        \item[2.] Random locations in space/time left out;
        
        \item[3.] All spatial  locations are missing on the last 10 time points.
       
    \end{itemize} 
    For more details on the simulation settings, see Section 3.2 of \cite{nag_2022}. In the competition, we have considered the following architecture for \textbf{Space-Time.DeepKriging}. The point predictions in all scenarios is the median quantile with $\tau = 0.5$. Through trial and error, we adjusted the number of nodes and layers to give reasonable performance. 

\begin{itemize}
    \item We have used $3$ layers of spatial basis functions with $G = 25,81,144$ and $3$ layers of temporal basis with $H = 10,15,45$ respectively. We had $297$ basis functions in total. 
    \item The neural network consists of $L=13$ layers with $M_i=100, i = 1,...,8$, $M_i=50, i = 9,...,12$ and $M_{13} = 1$ as the final layer.
    \item We initialized the weight matrices by taking samples from normal distribution. We used L1L2 regularization in the first two layers, and chose $0.001$ as the learning rate. Here L1 \citep{vidaurre2013survey} and L2 \citep{cortes2012l2} regularization carries the same meaning as the LASSO and ridge regression penalization on parameters to prevent overfitting. We have used a combination of both the regularization techniques. 
    \item Our network architecture remained the same for all the datasets. However, we selected different values of the hyper-parameters such as the number of epochs and batch size based on convergence.

    
\end{itemize}

For the forecast tasks in the competition we only used \textbf{QLSTM} with $P=1$ and $50$ hidden nodes.

Our model performed the best among seven competing teams. Approaches considered by other teams include Vecchia's approximation by \textbf{GpGp} \citep{guinness2021gpgp} (ranked the second) and block-composite likelihood  by \textbf{RESSTE} (ranked the third). \textbf{RESSTE} also used the R-package developed by \textbf{GpGp} \citep{guinness2021gpgp} for their implementation. More details on these methods can be found in \cite{sun2012geostatistics}. We got an overall root mean squared error of \textbf{0.25} for sub-compatition 2a ($22$\% improvement over the second best) and \textbf{0.26} for sub-compatition 2b ($9$\% improvement over the second best). Details of the RMSPE corresponding to each dataset and also the evaluation procedure can be found in \cite{nag_2022} and the supplementary materials therein.

\subsection{Nonstationary space-time simulation}

We simulate a nonstationary space-time data over $100$ spatial and $500$ temporal locations. We first generate a zero-mean stationary Gaussian process with the space-time Mat\'ern covariance function as discussed in \eqref{eq:st-cov} with the same parameter settings as competition data described in Section \ref{sec:competition}. Unlike the competition data, we also introduce mean nonstationarity in the data by adding 
$$
\mu(t) = 2\sin\left[15 \left(\frac{t}{1000} - 0.9\right) \right] \cos \left[-37\left(\frac{t}{1000} - 0.9\right)^4\right] + \frac{\left(\frac{t}{1000} - 0.9\right)}{2}.
$$

To compare the interpolation and forecast predictions of our proposed approach with other established approaches for large datasets we compared our results with \textbf{GpGp}~\citep{guinness2021gpgp} which uses Vecchia approximation for large scale spatial interpolation and forecasting. To make the \textbf{GpGp} model more competitive we fitted it on the stationary data instead and added the true nonstationary mean to obtain the final prediction. The number of nearest neighbors is set to be 10 and 30 sequentially in Vecchia approximation. Table \ref{tab:simulation_prediction} shows the comparison results for a 10-fold cross validation on the data. Here we trained both the models on $45,000$ space-time locations and tested on $5000$ locations. It is clear that \textbf{Space-Time.DeepKriging} outperforms \textbf{GpGp} on mean square prediction error (MSPE).

\begin{table}[htp!]
    \centering
    \caption{Average MSPE of prediction for simulated data. Here SE stands for standard error of the predictions.}
    \vspace{2mm}
        \begin{tabular}{||c | c c ||} 
        \hline
        Models & MSPE & SE \\ [0.5ex] 
         \hline\hline
         \textbf{Space-Time.DeepKriging} & \textbf{0.167} & 0.073 \\
         \textbf{GpGp} & 0.746 & 0.288 \\
         \hline
         \end{tabular}
    \label{tab:simulation_prediction}
\end{table}

Figure 3 provides the visual representations of interpolation over three time points 250, 350 and 450. The last column of the plot shows the length of the $90\%$ prediction interval for the predictions of \textbf{Space-Time.DeepKriging}. Note that, to obtain the prediction intervals here we implement the quantile loss as discussed in \eqref{eq:loss_LSTM} in place of MSE loss to optimize the parameters for \textbf{Space-Time.DeepKriging}. 

\begin{figure}[htp!]
     \centering
     \includegraphics[width=0.3\textwidth]{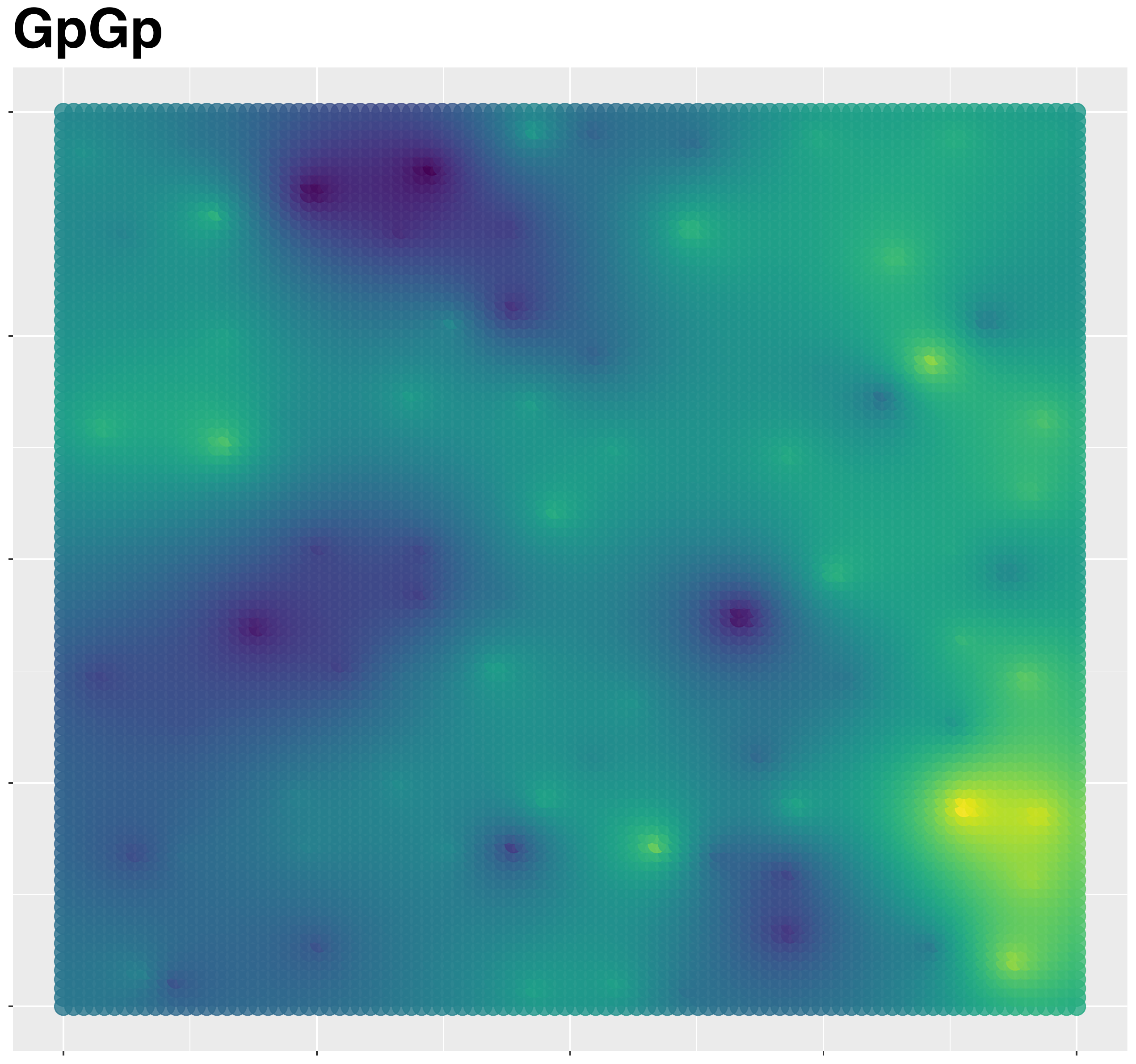} 
     \includegraphics[width=0.32\textwidth]{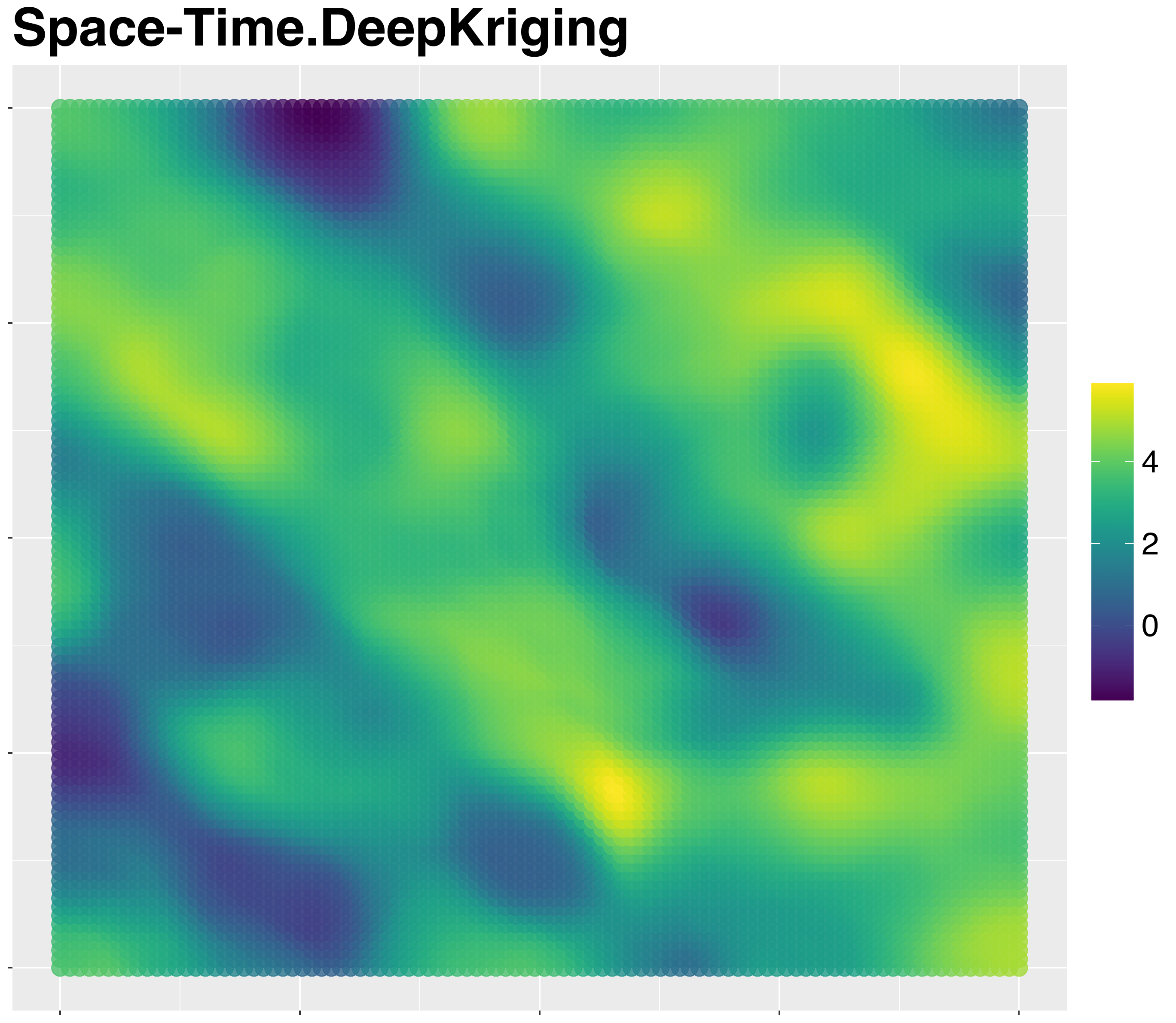} 
     \includegraphics[width=0.3\textwidth,height=4.6cm]
     {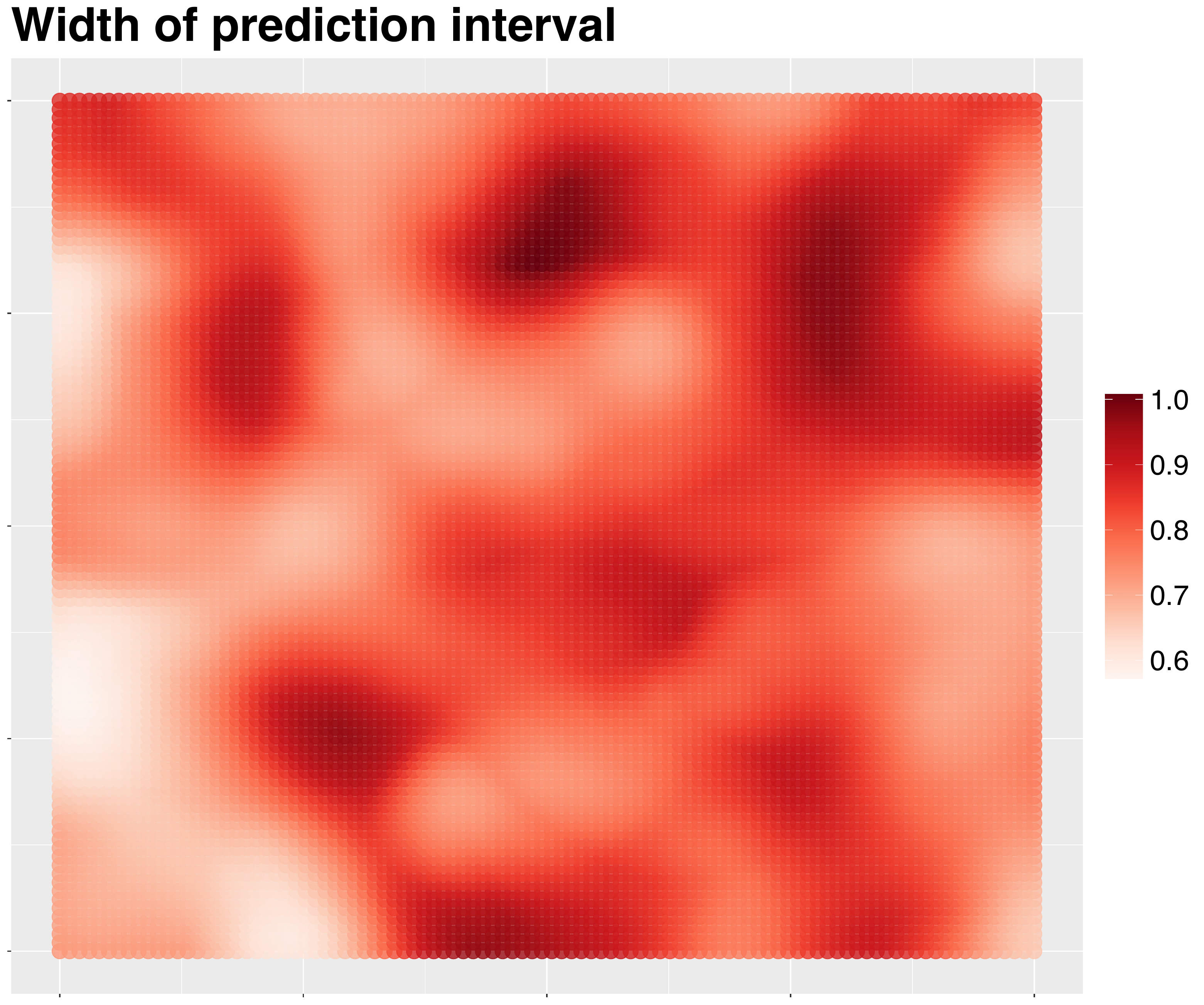} 
     
     \vspace{5mm}
     \includegraphics[width=0.3\textwidth]{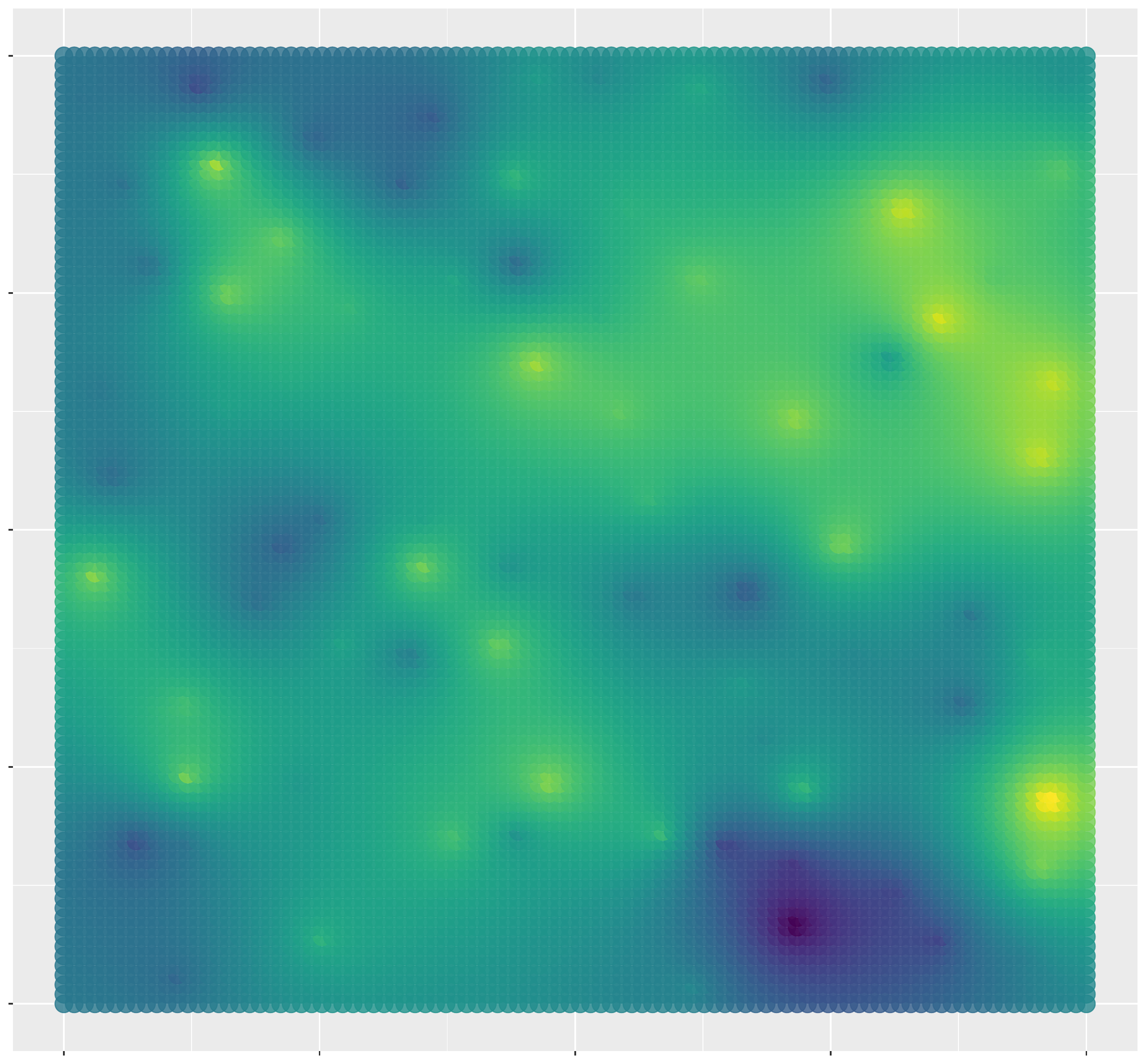} 
     \includegraphics[width=0.32\textwidth]{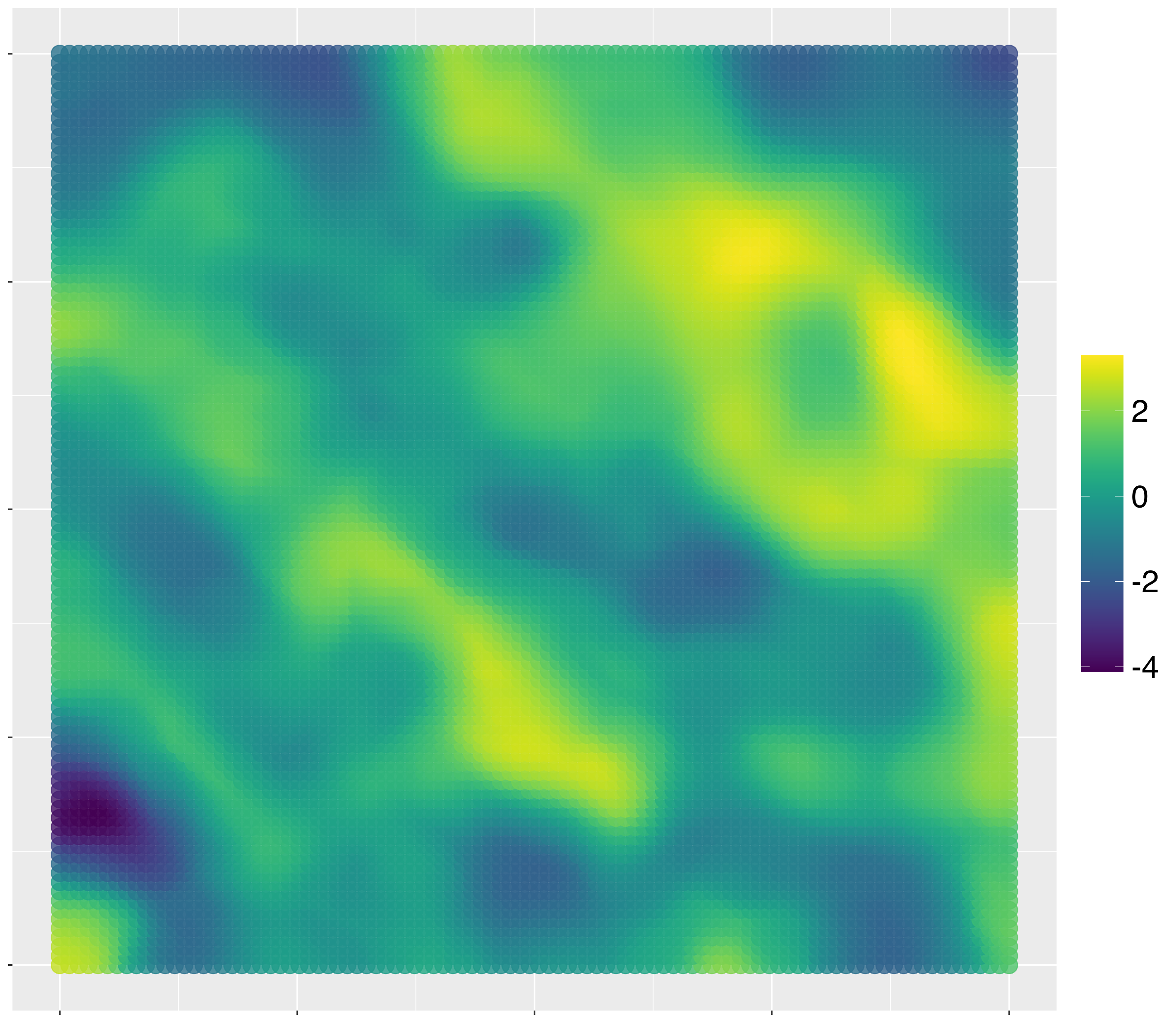} 
     \includegraphics[width=0.3\textwidth]{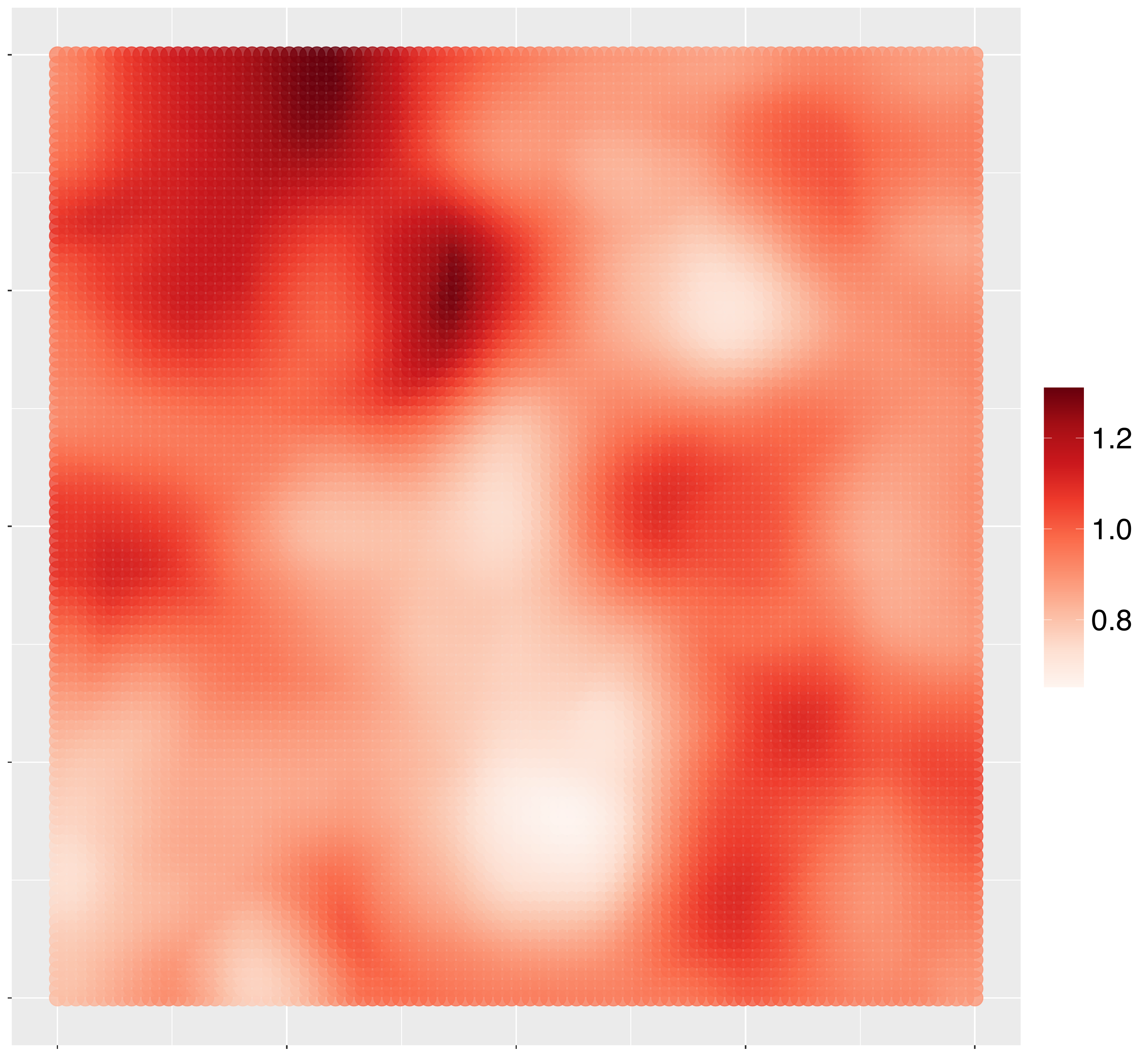}
     
     \vspace{5mm}
     
     \includegraphics[width=0.3\textwidth]{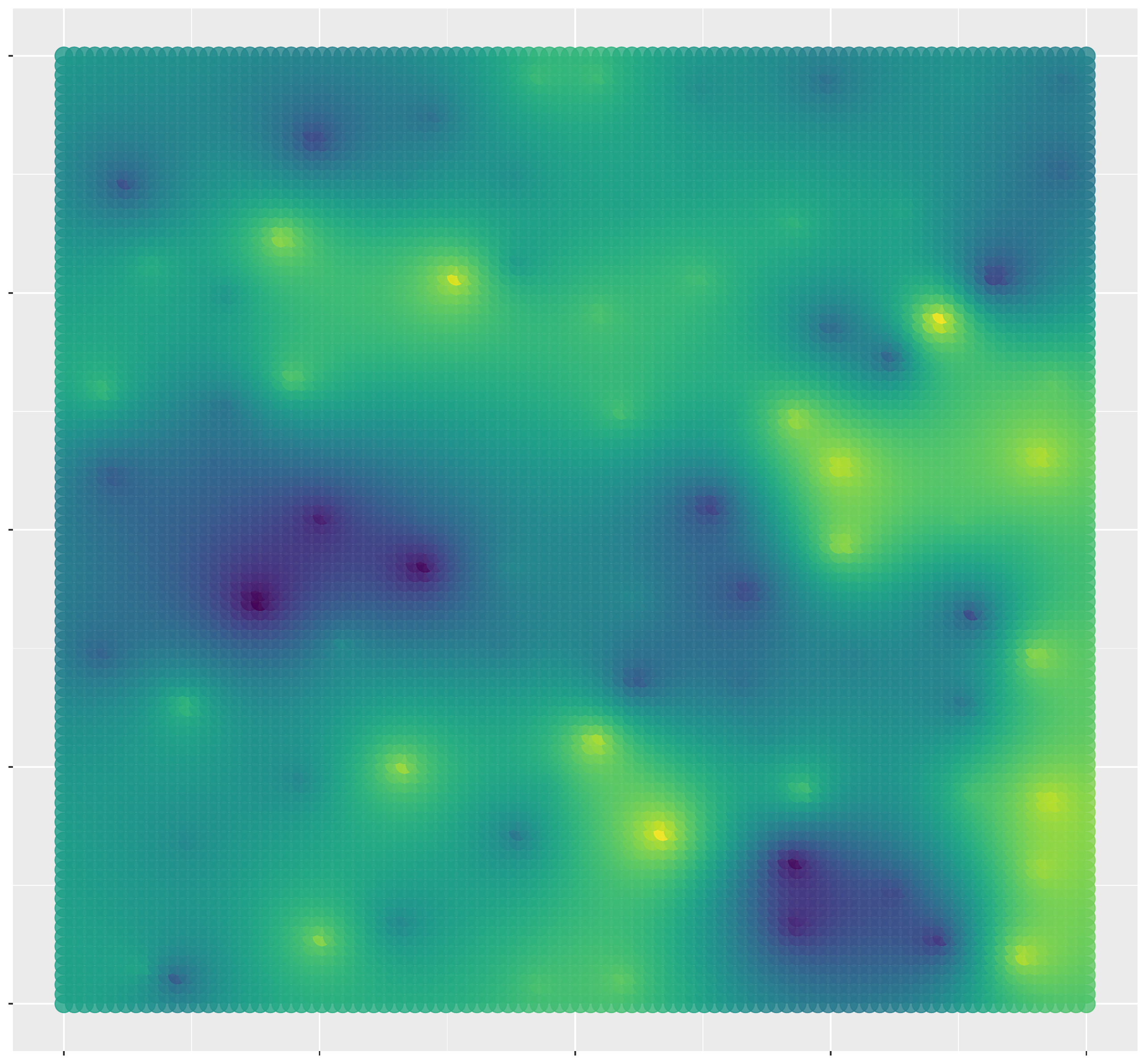} 
     \includegraphics[width=0.32\textwidth]{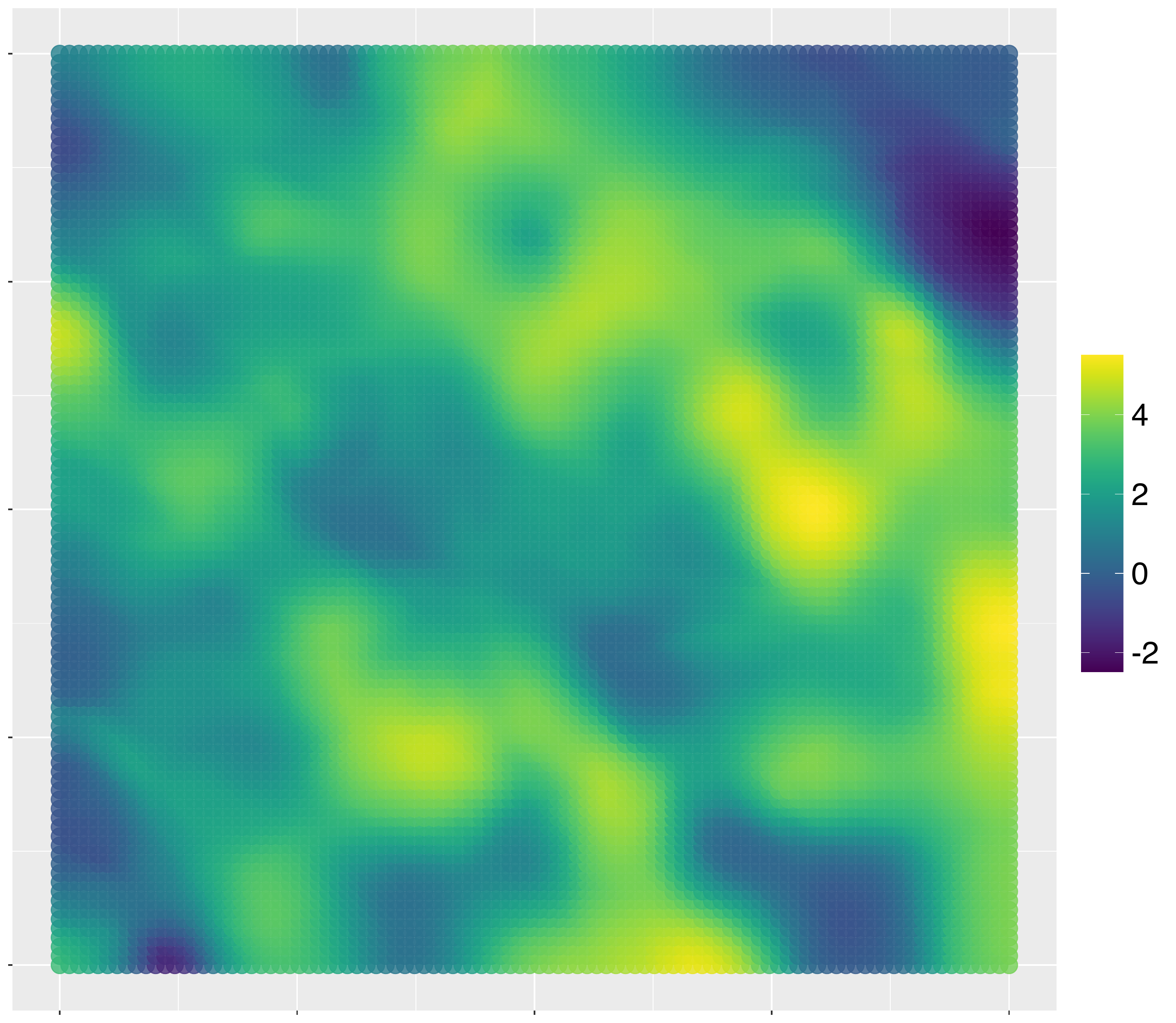} 
     \includegraphics[width=0.3\textwidth]{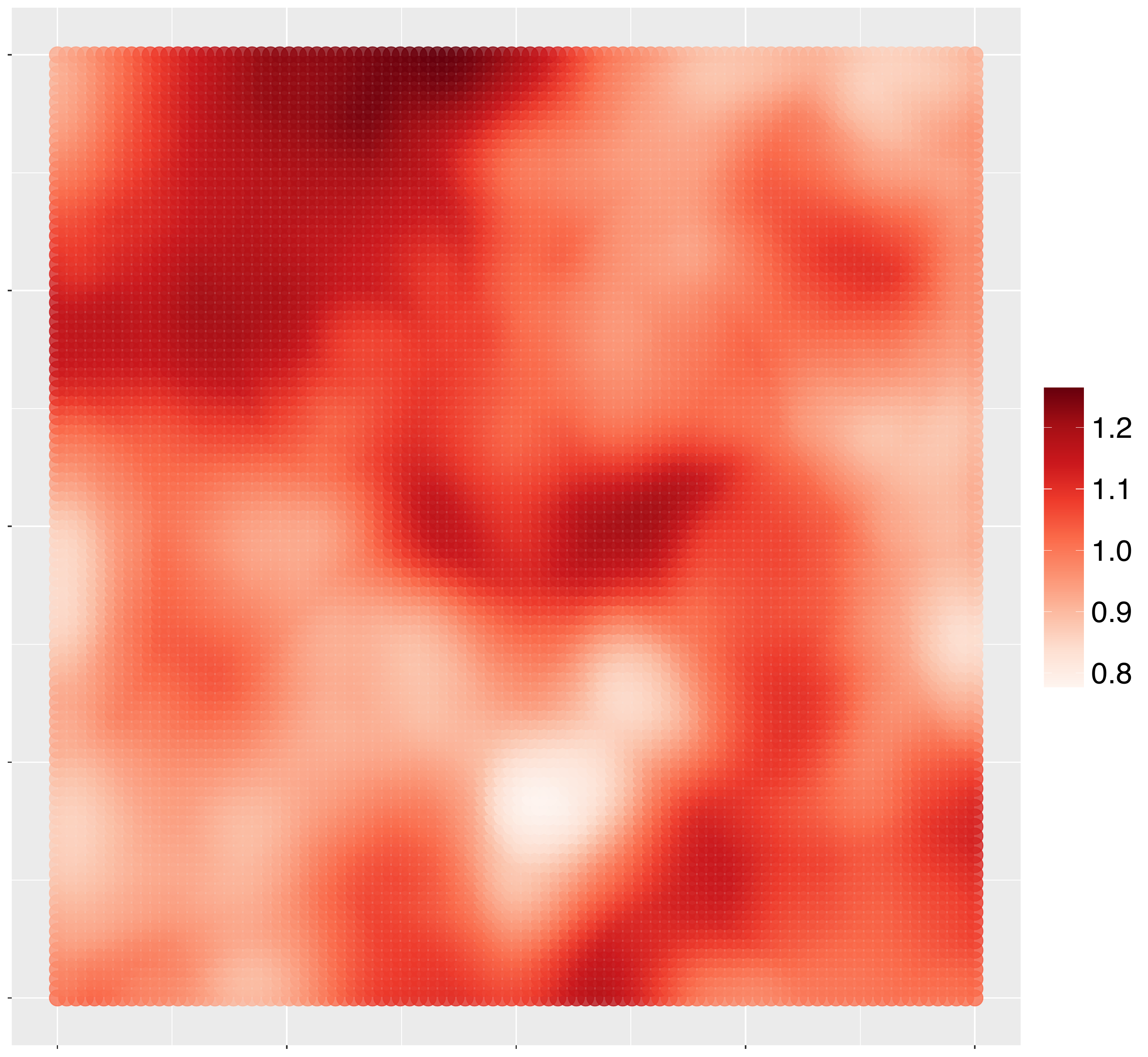}
     \label{fig:simulation}
    \caption{Interpolation over the $[0,1] \times [0,1]$ square grid with \textbf{GpGp} and \textbf{Space-Time.DeepKriging}. Column 1 shows the interpolation for time points 250,350 and 450 with \textbf{GpGp}, Column 2  shows the interpolation with \textbf{Space-Time.DeepKriging} and column 3 provides the width of the 90\% prediction intervals obtained through \textbf{Space-Time.DeepKriging}.}
\end{figure}

To justify the efficiency of our proposed probabilistic forecasting we have compared the performance of our forecast and 90\% forecasting interval with \textbf{GpGp} and autoregressive integrated moving average (ARIMA). We train each models on $100$ spatial locations with $495$ time points and leave the last $5$ time points for testing. For example, we train the \textbf{Space-Time.DeepKriging} untill time points $495$ and then train \textbf{QConvLSTM} to forecast the observations at each location for the last 5 time points. We use \pkg{auto.arima} function from the package \pkg{forecast} for fitting the ARIMA models separately for each location for 100 locations. Table \ref{tab:simulation_forecast} shows the average MSPE, mean prediction interval width (MPIW) and prediction interval coverage (COV) for all the competing models. Note that, package \textbf{GpGp} did not have options to obtain the prediction variance through which prediction interval can be calculated. Clearly, it can be observed that \textbf{QConvLSTM} performs better than other competing models based on MSPE and MPIW. Moreover, it can be observed that \textbf{QConvLSTM} has about $31\%$ improvement over \textbf{QLSTM}, demonstrating that incorporating local spatial information increases forecast accuracy.
\begin{table}[h!]
    \centering
    \caption{Average mean square prediction error (MSPE), mean prediction interval width (MPIW) and and 90\% interval coverage (COV) of forecast for simulated data.}
    \vspace{2mm}
        \begin{tabular}{||c | c c c c c||} 
        \hline
        Models & MSPE & SE & MPIW & SE & COV(\%)\\ [0.5ex] 
         \hline\hline
         \textbf{QConvLSTM} & 0.267 & 0.219 & 1.462 & 0.126 & 90.39 \\
         \textbf{ARIMA} & 0.277 & 0.278 & 2.262 & 0.082 & 90.72 \\
         \textbf{QLSTM} & 0.392 & 0.523 & 1.558 & 0.316 & 89.94 \\
          \textbf{GpGp} & 0.839 & 0.358 & - & - & -\\
         \hline
         \end{tabular}
    \label{tab:simulation_forecast}
\end{table}

\begin{figure}[h!]
    \centering
    \subfigure[]{
     \centering
     \includegraphics[width=\textwidth]{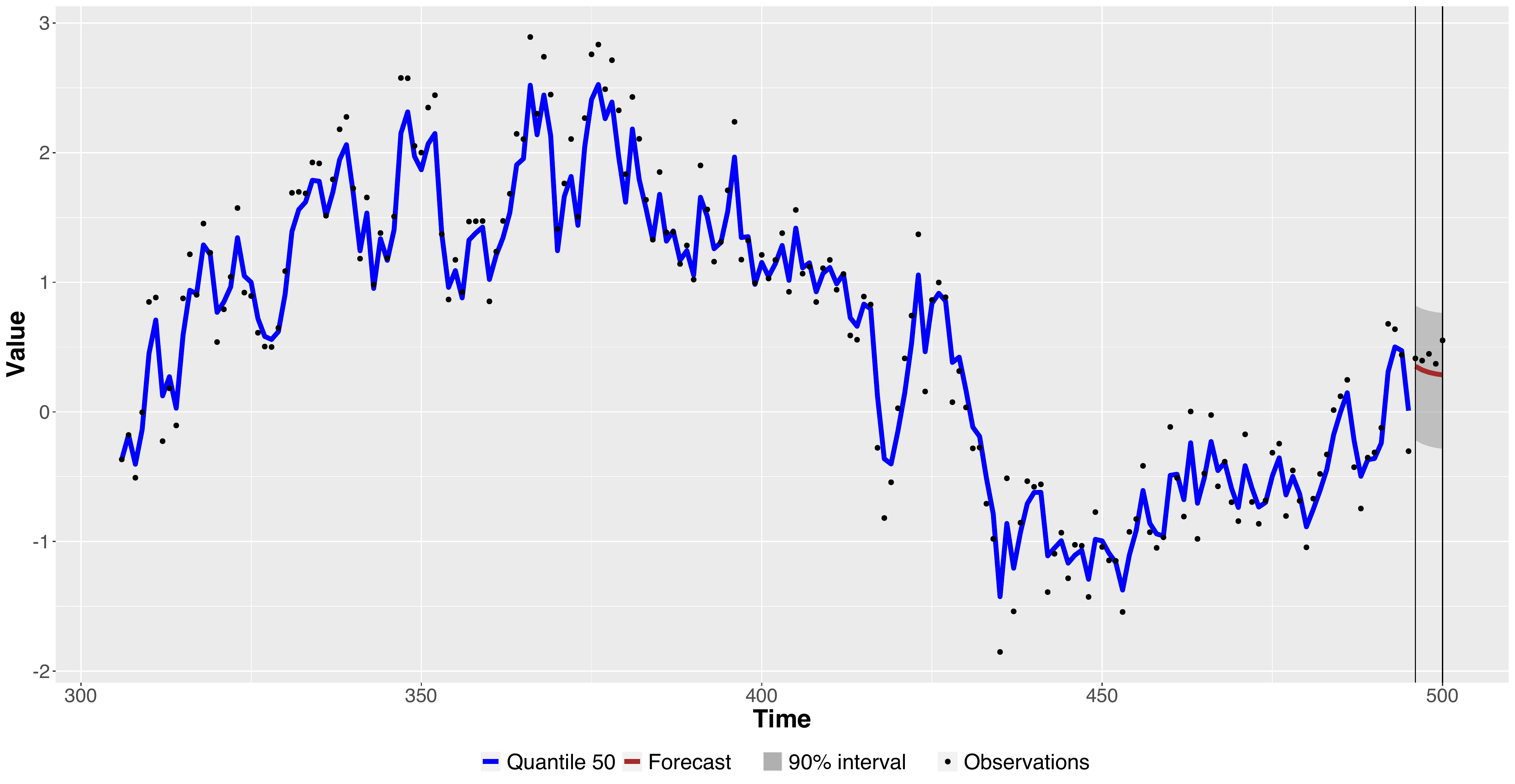} 
    }
    \subfigure[]{
     \centering
     \includegraphics[width=\textwidth]{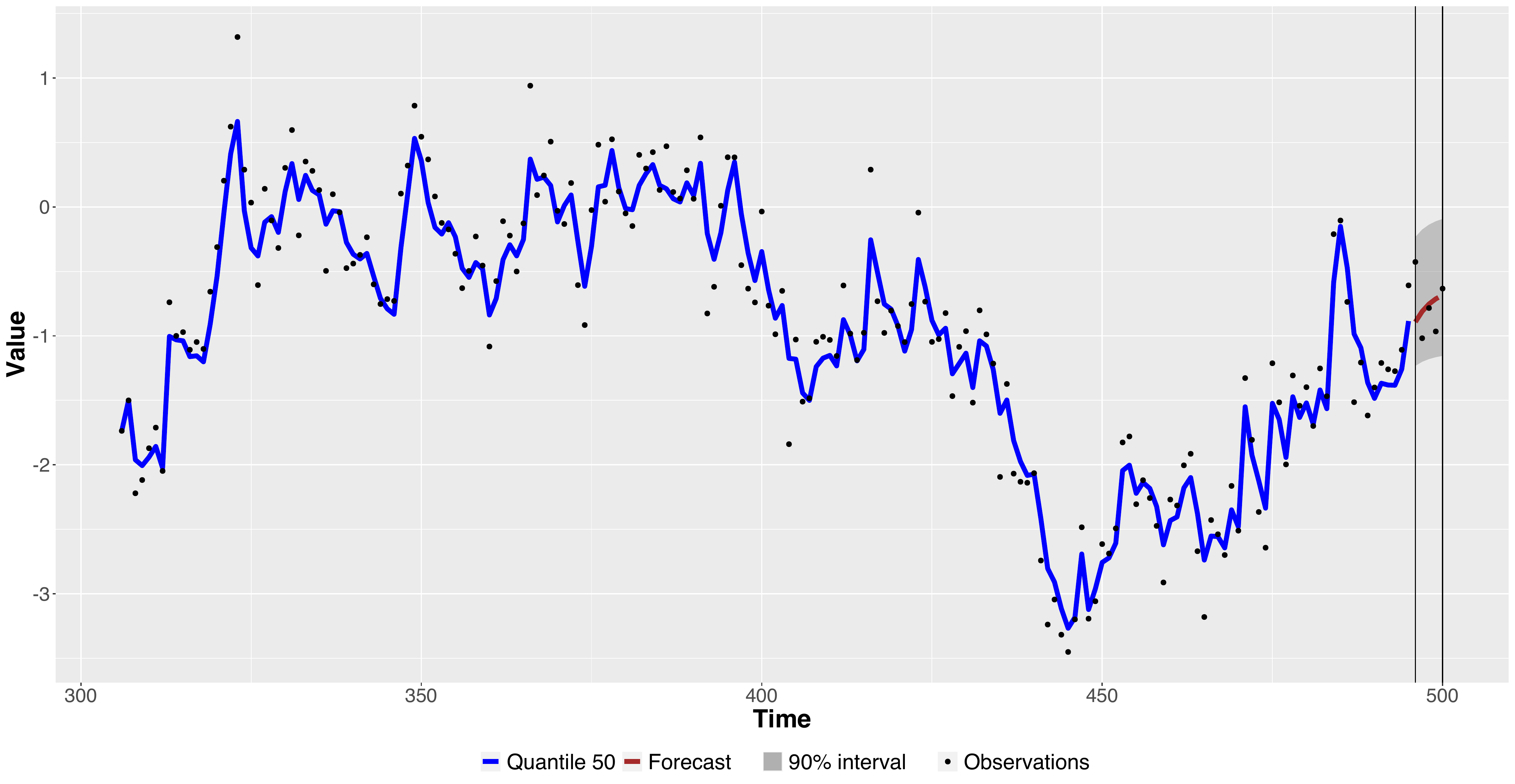}
    }
    \caption{(a) Forecast for location 1, (b) Forecast for location 2 for the simulated data.}
    \label{fig:forecast_competition}
\end{figure}

Figure~\ref{fig:forecast_competition} shows the time series plot for two spatial locations ($0.0161, 0.0042$, location 1) on bottom left and ($0.7561, $ $ 0.7744$, location 2)on top right. The blue line represents the 50-th quantile obtained by training \textbf{QConvLSTM}. The brown line with gray shade towards the end of the time series showcases 50-th forecast quantile and the $90\%$ forecasting interval for the last 5 time points. Note that, here we have only shown the last $200$ time points of the series.


\section{Application to \texorpdfstring{$PM_{2.5}$ } \  data}\label{sec:prediction3}

The air pollutant $PM_{2.5}$, or fine particulate matter smaller than $2.5\mu m$ is known to have adverse health effects. The World Health Organization (2013) summarizes its harmful effects including respiratory disease \citep{peng2009emergency} and myocardial infarction \citep{peters2001increased}. To evaluate the effects of $PM_{2.5}$ exposure, a high-resolution map of $PM_{2.5}$ concentration is required. The most accurate estimation of $PM_{2.5}$ concentration at a specific time and location is provided by measurements from monitoring networks.
However, data from monitoring sites are sparse and thus interpolating $PM_{2.5}$ concentration using data from monitoring networks is an important application for spatio-temporal methods.

The US EPA maintains $PM_{2.5}$ measurements for over two decades in the United States.   
We analyze monthly data from 1999 to 2022. The dataset contains a total of $286$ months. The monitoring sites are irregularly spaced across the US. On average, there are 1,900 spatial observations per month. 
\begin{figure}[htp!]
    \centering
    \subfigure[]{
     \begin{minipage}{\linewidth}
     \centering
     \includegraphics[width=0.327\textwidth]{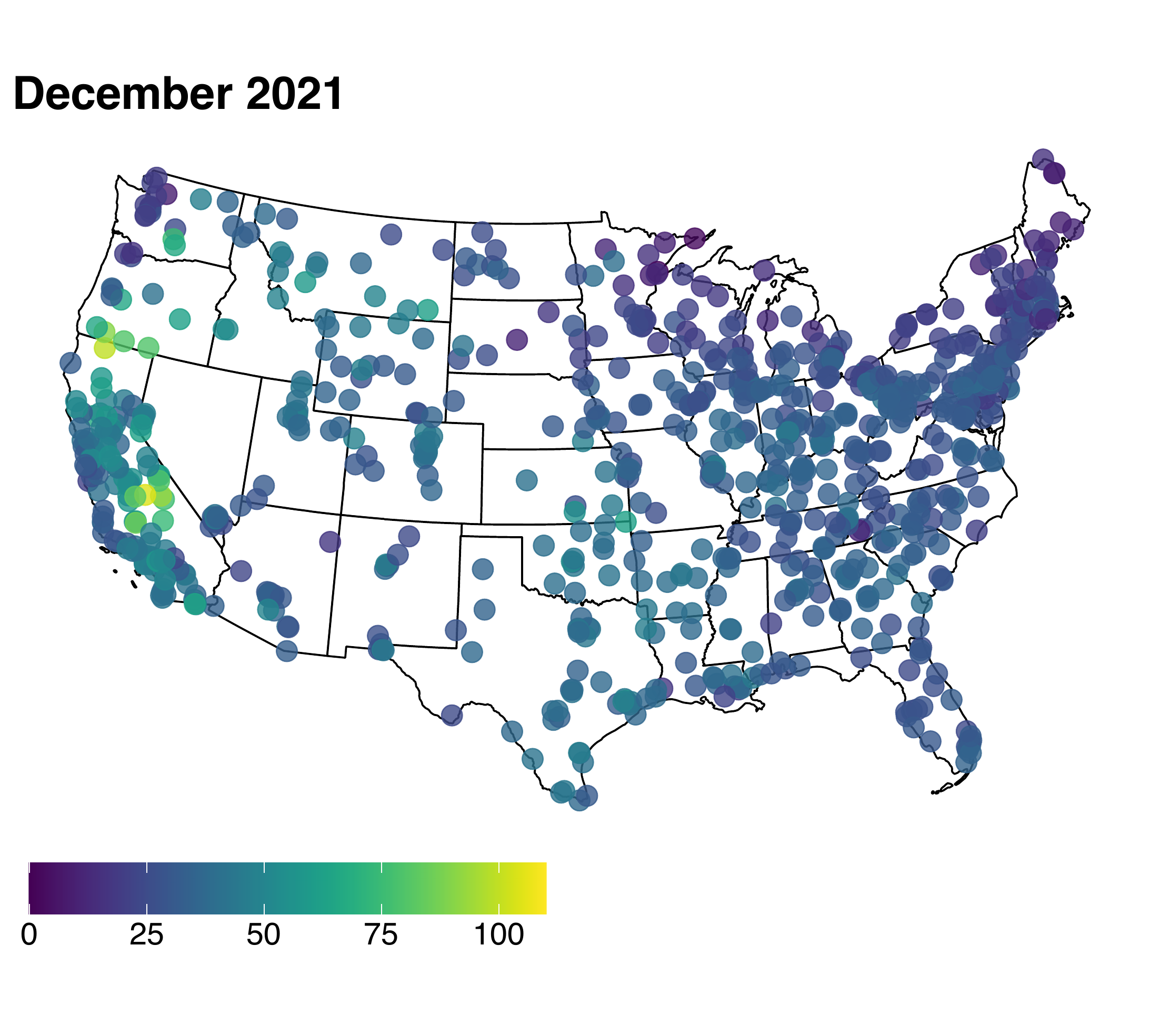} 
     \includegraphics[width=0.327\textwidth]{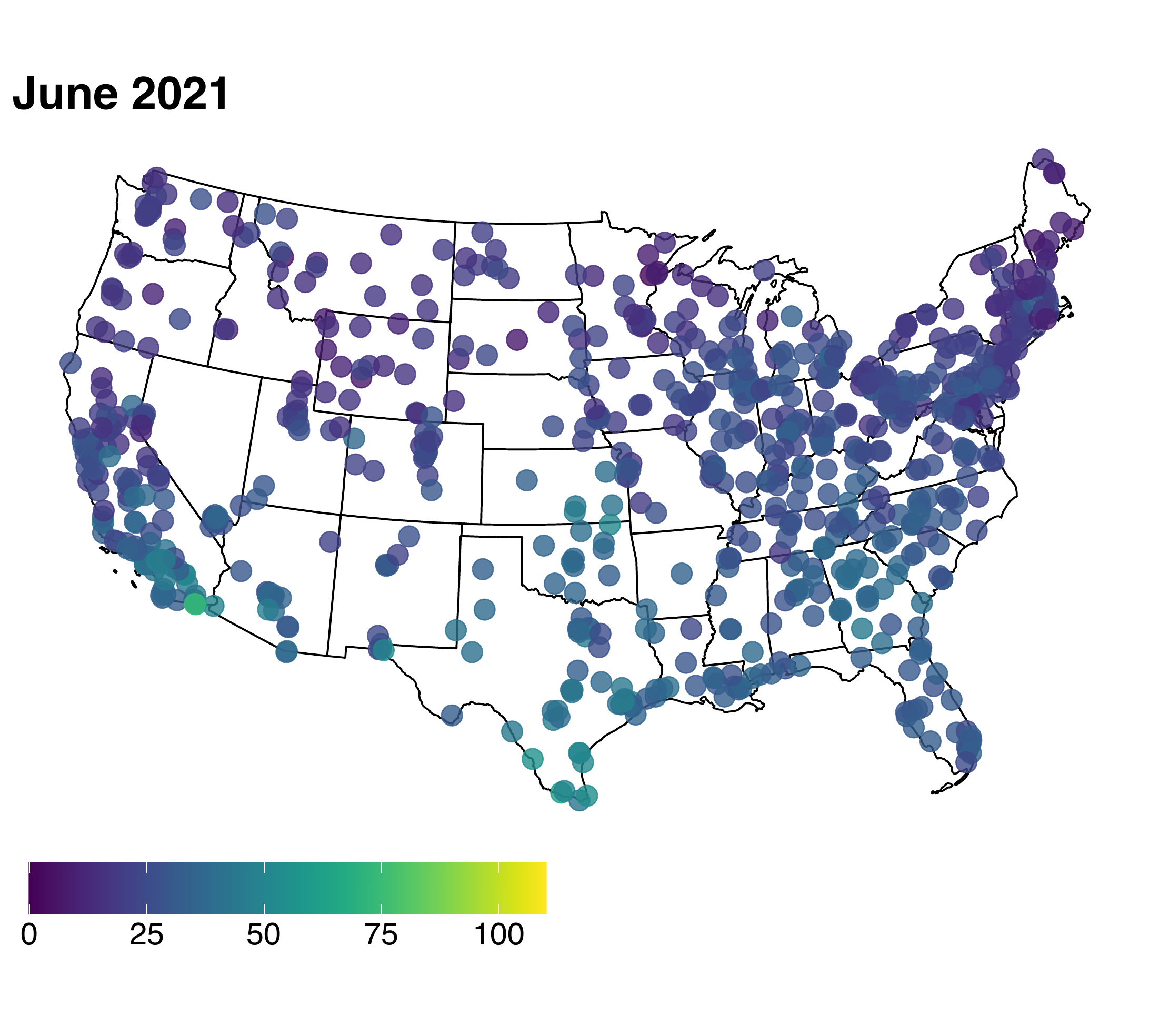} 
     \includegraphics[width=0.327\textwidth]{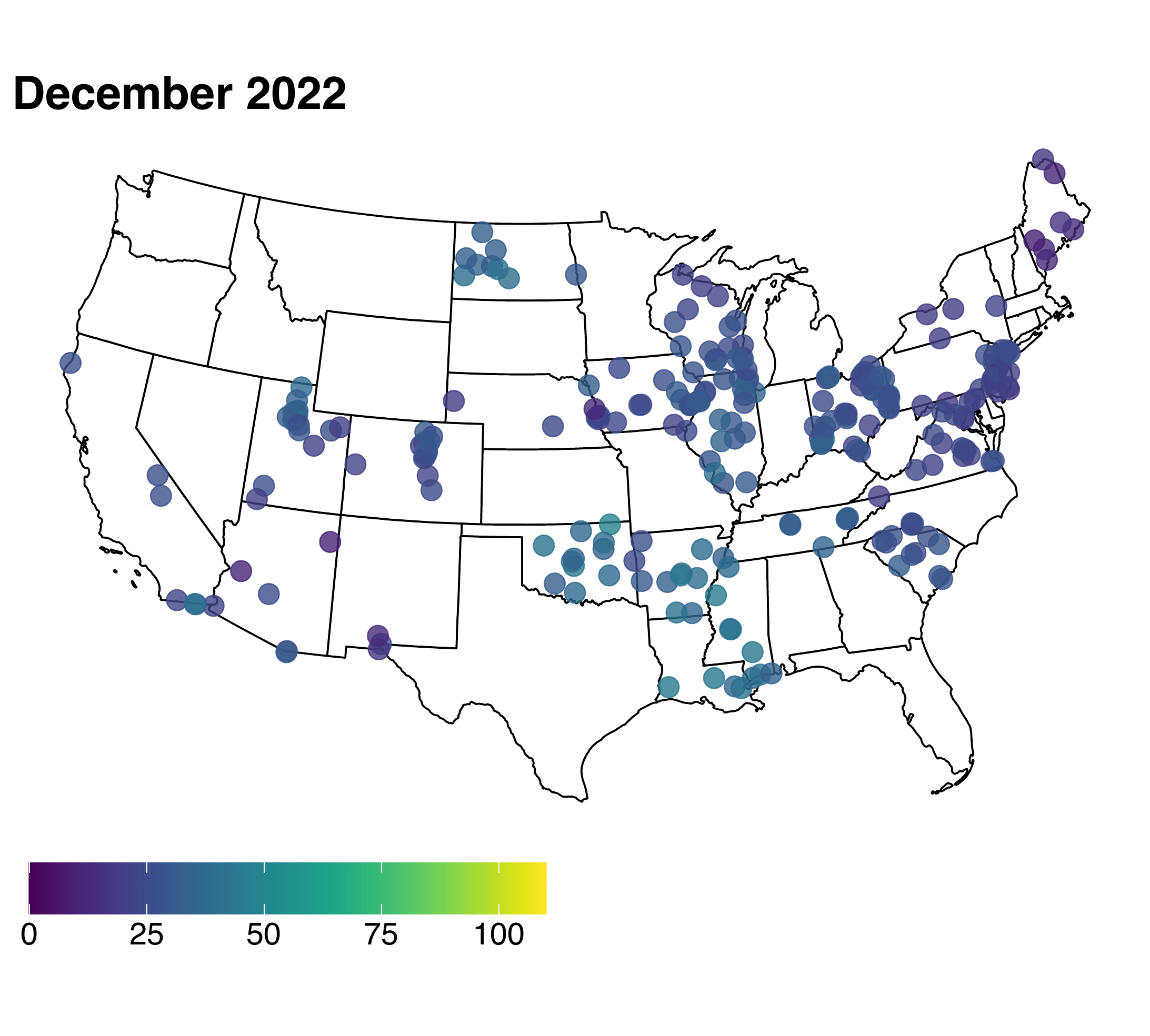} 
      \vspace{-5mm}
      \end{minipage}
      \label{fig:real_example}
    }
    \vspace{-2mm}
    \subfigure[]{
     \begin{minipage}{\linewidth}
     \centering
     \includegraphics[width=0.327\textwidth]{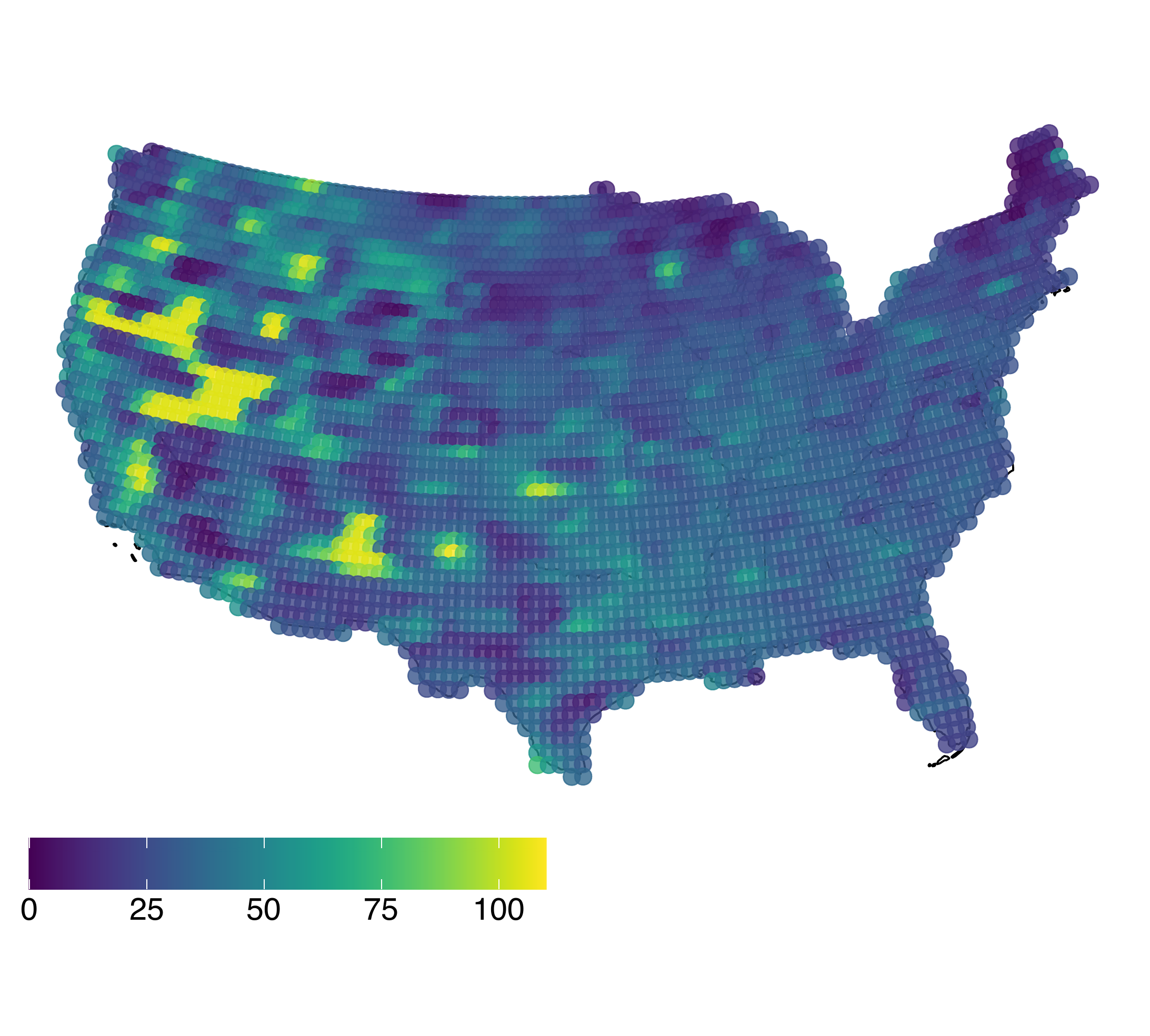}
     \includegraphics[width=0.327\textwidth]{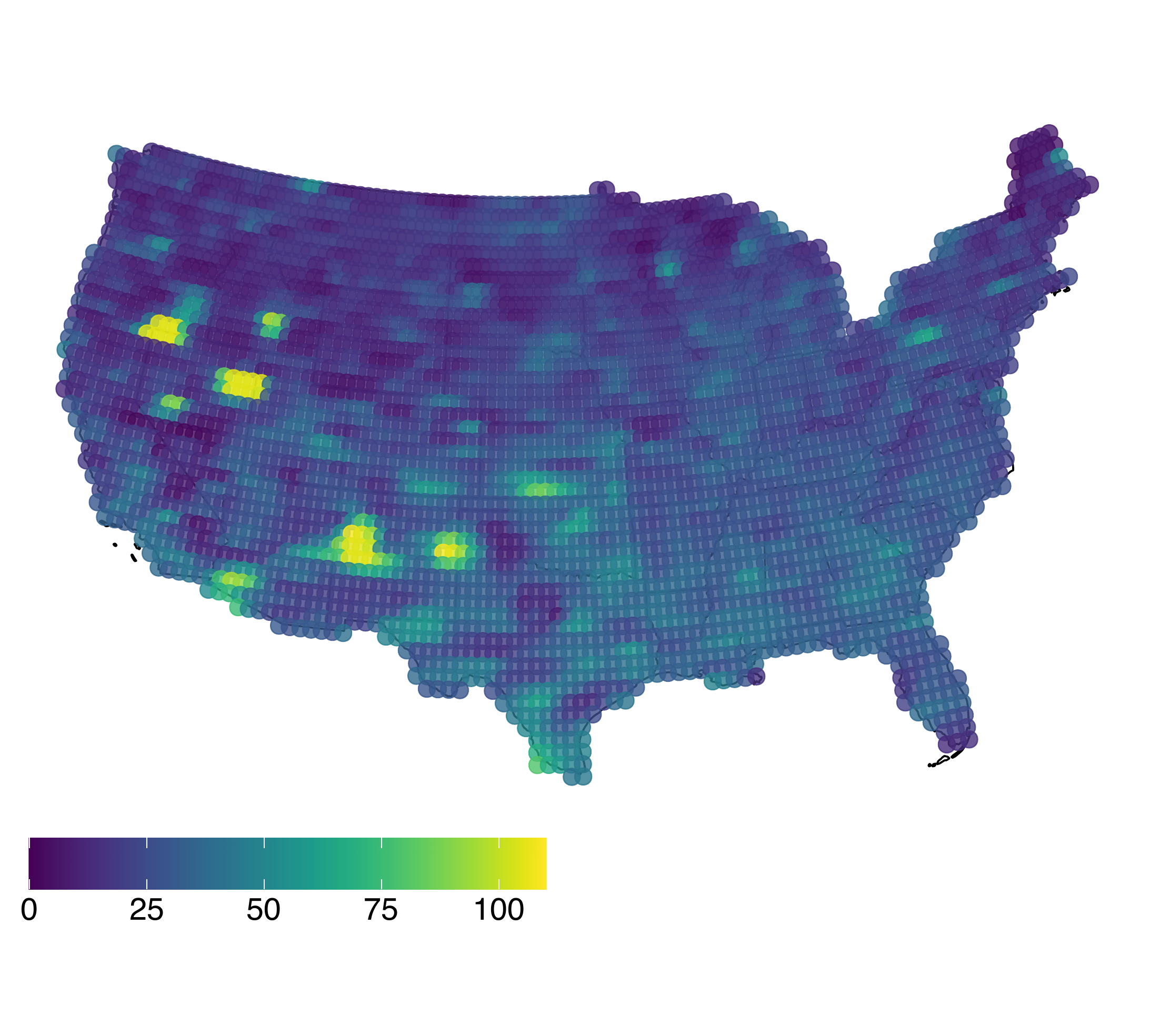}
     \includegraphics[width=0.327\textwidth]{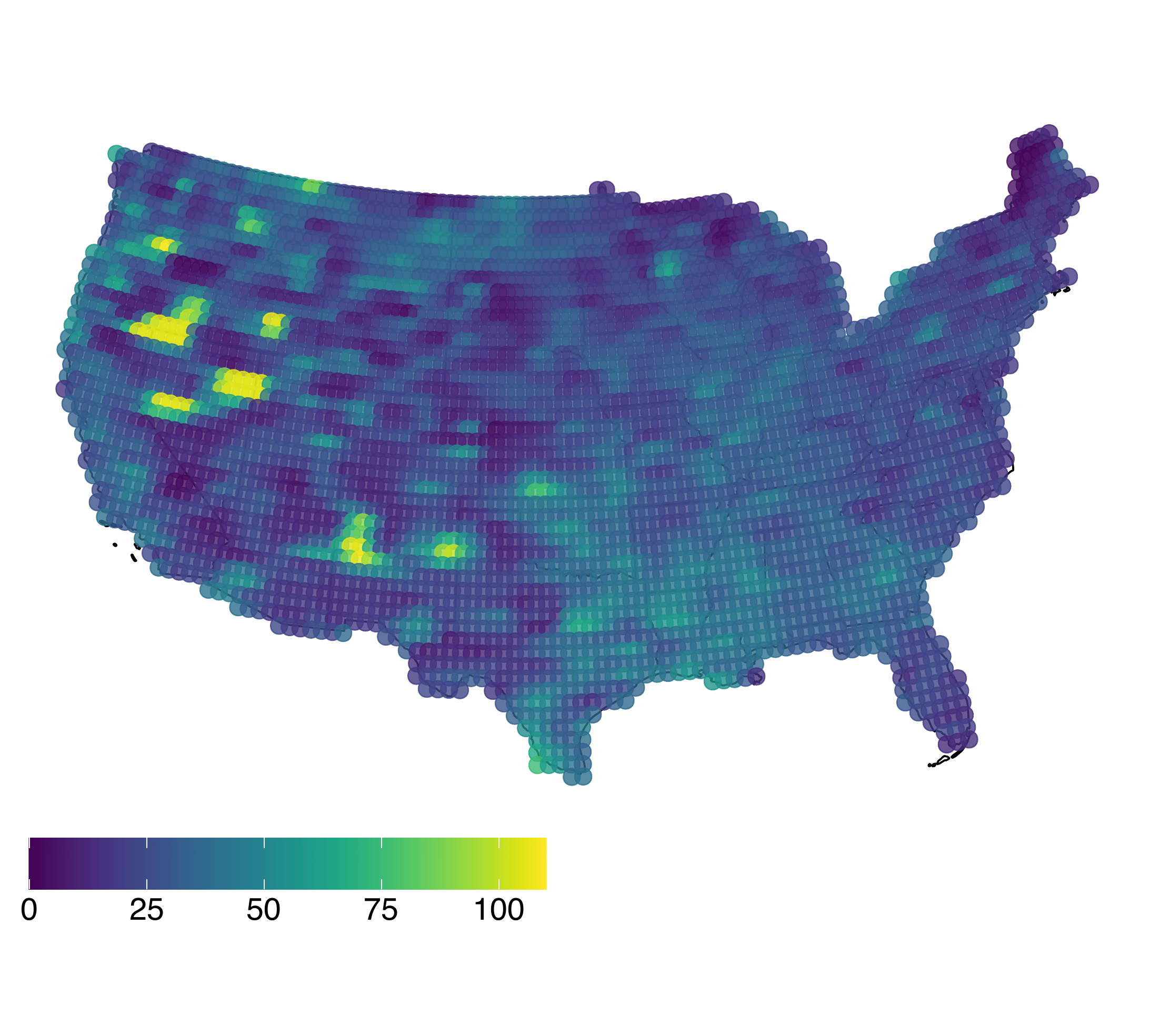}
      \vspace{-5mm}
      \end{minipage}
      \label{fig:real_interpolation}
    }

    \vspace{-2mm}
    \subfigure[]{
     \begin{minipage}{\linewidth}
     \centering
     \includegraphics[width=0.327\textwidth]{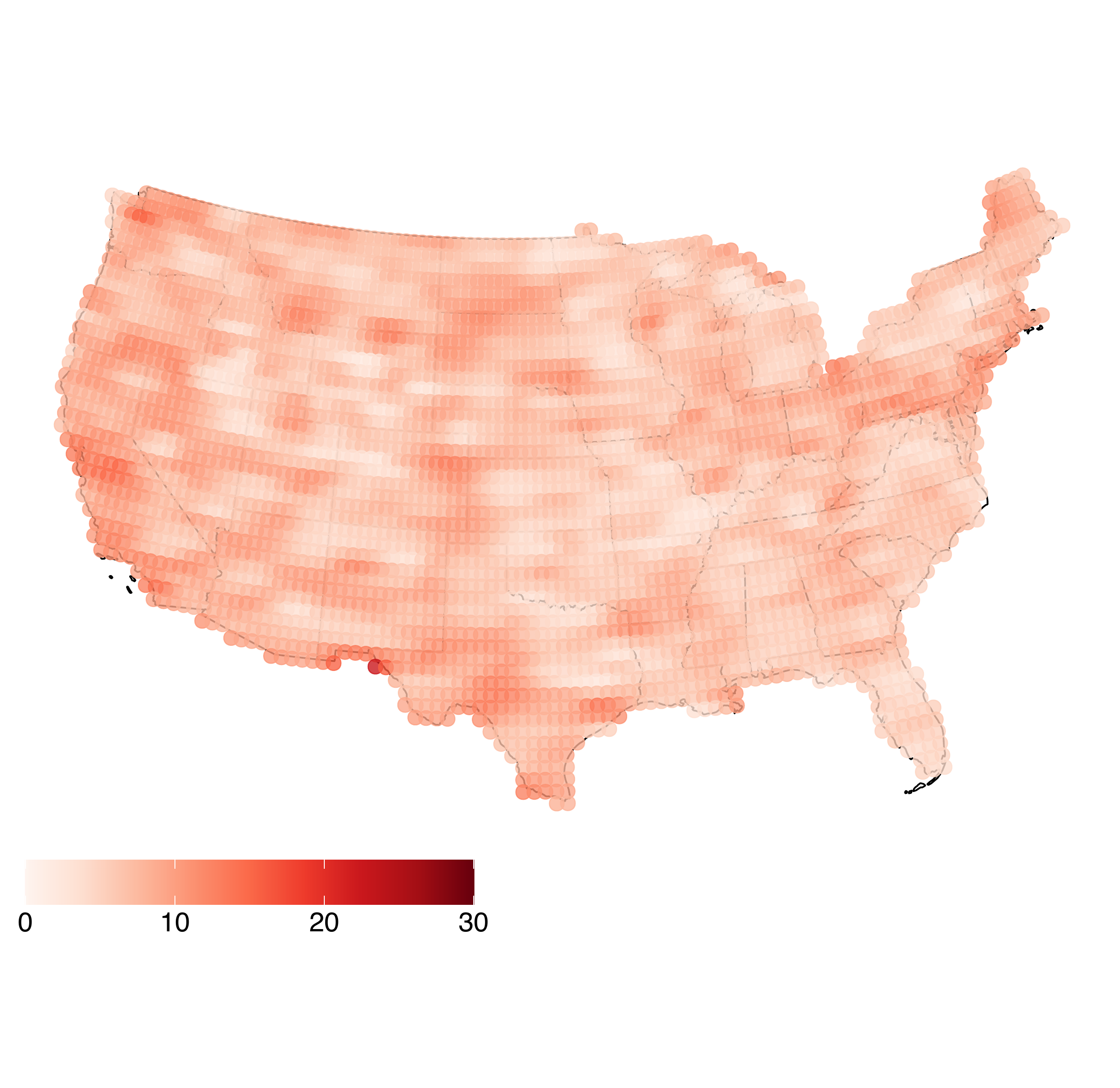}
     \includegraphics[width=0.327\textwidth]{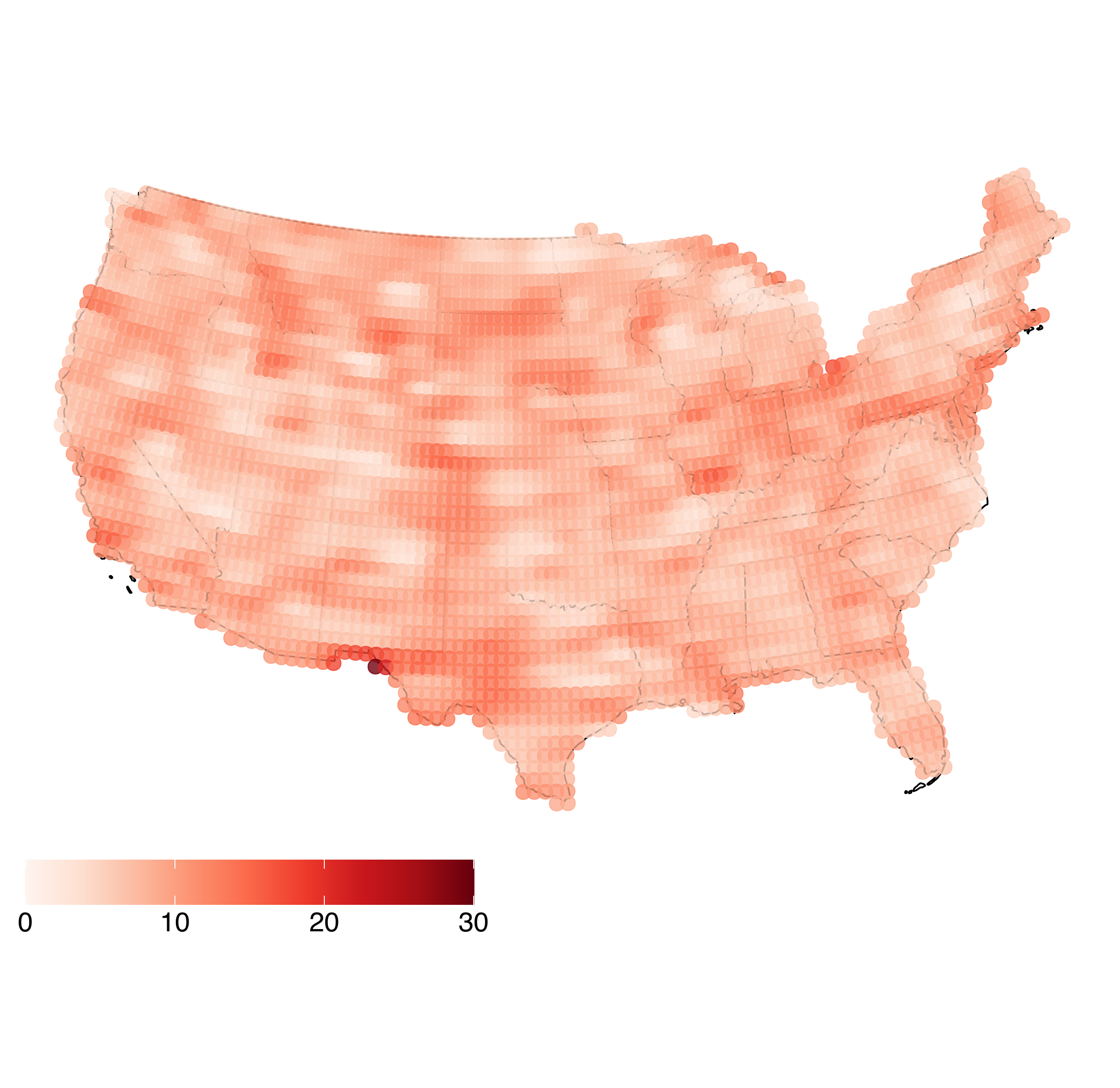}
     \includegraphics[width=0.327\textwidth]{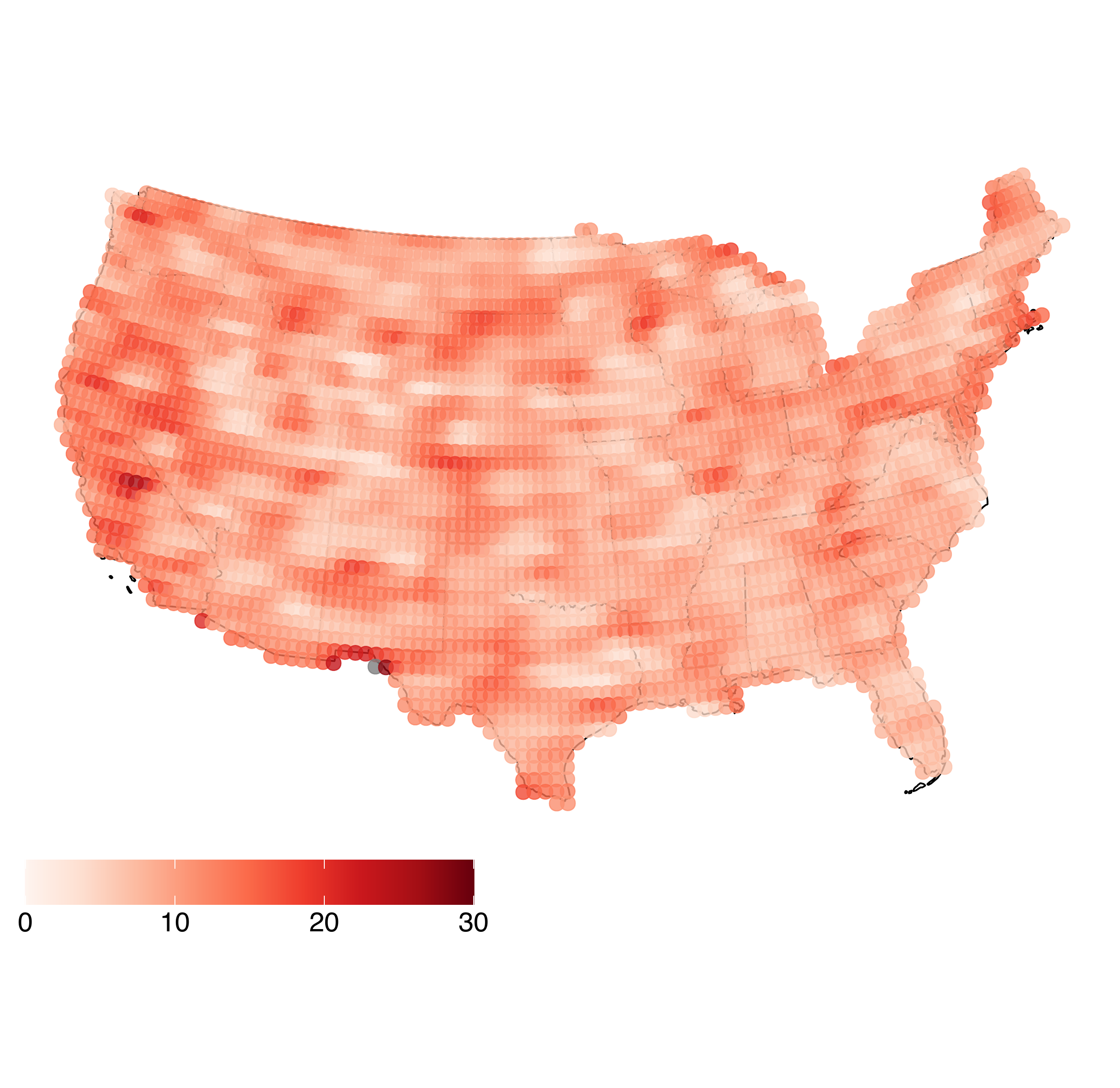}
      \vspace{-5mm}
      \end{minipage}
      \label{fig:real_uncertainty}
    }
    \caption{(a) $PM_{2.5}$ (in Micrograms/cubic meter) for three selected months,  (b) Space-time interpolation for these three months using \textbf{Space-Time.DeepKriging}, and (c) width of the 90\% prediction interval.}
    \label{fig:real_all}
\end{figure}



We compare different methods using cross-validation. For training, we have randomly chosen $N=200,000$ of the $261,487$ space-time locations.
We fit \textbf{Space-Time.DeepKriging} on the dataset with the same framework discussed in Section \ref{sec:prediction2}. Figure \ref{fig:real_interpolation} shows the interpolation results. Clearly, \textbf{Space-Time.DeepKriging} has successfully captured the spatial patterns in the data. There are few instances where the interpolation predicts high values in certain locations where all the nearby locations have low values, but these regions have large $PM_{2.5}$ at prior months. From Table \ref{tab:pm_interp} we can see that \textbf{Space-Time.DeepKriging} has a smaller mean square prediction error (MSPE) than \textbf{GpGp}. We have also provided the width of the $90\%$ prediction intervals for each of the interpolated sites in Figure \ref{fig:real_uncertainty}. There were very few observations for December 2022 and thus the prediction intervals are wider for this month.

\begin{table}[htp!]
    \centering
    \caption{Average mean square prediction error (MSPE) of the $PM_{2.5}$ interpolation and average standard error (SE).}
    \vspace{2mm}
        \begin{tabular}{||c | c c ||} 
        \hline
        Models & MSPE & SE \\ [0.5ex] 
         \hline\hline
         \textbf{Space-Time.DeepKriging} & \textbf{21.35} & 11.13 \\
         \textbf{GpGp} & 63.72 & 13.56 \\
         \hline
         \end{tabular}
    \label{tab:pm_interp}
\end{table}

\begin{figure}[h!]
    \centering
    \subfigure[]{
     \begin{minipage}{\linewidth}
     \centering
     \includegraphics[width=0.3\textwidth]{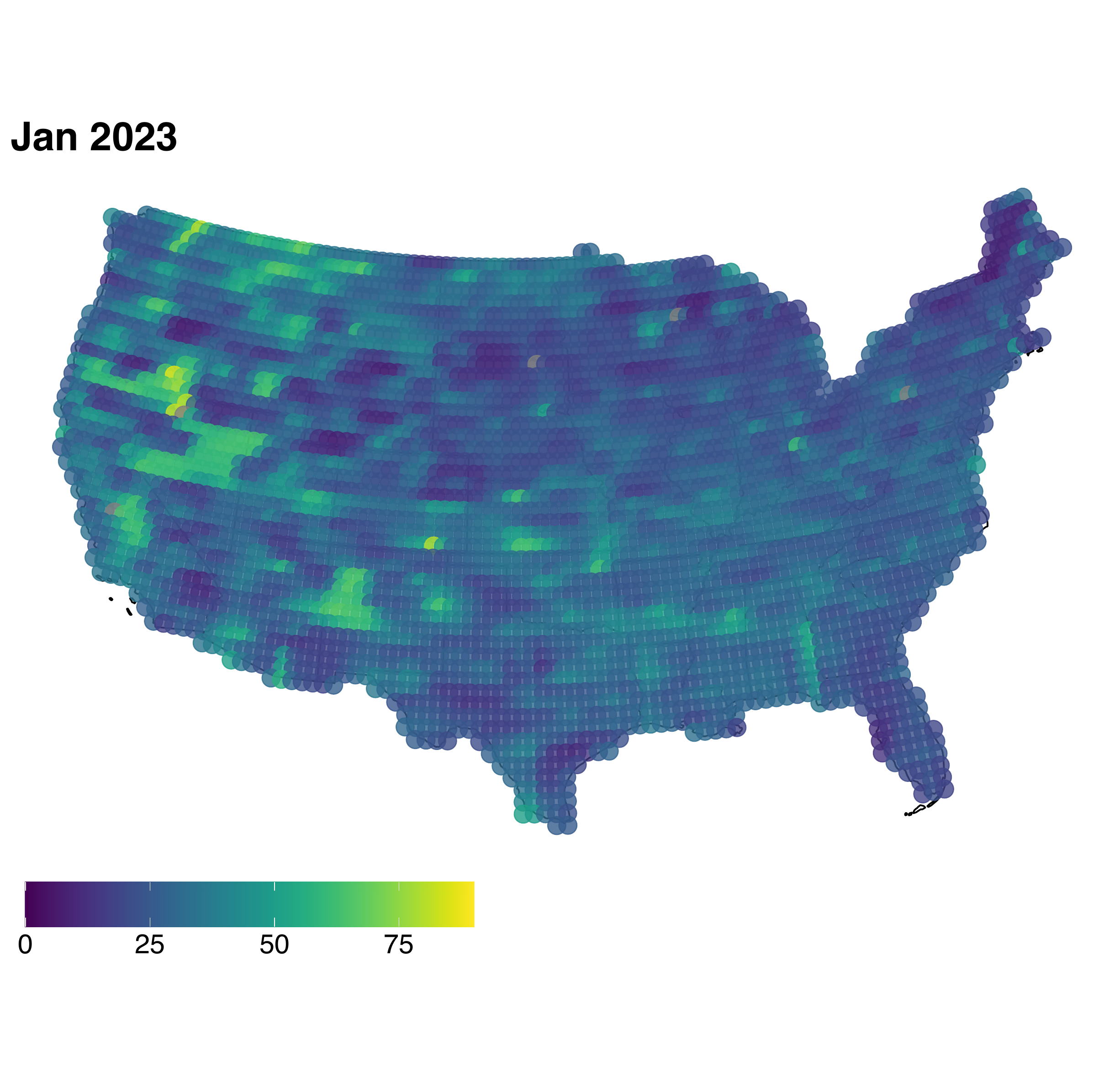} 
     \includegraphics[width=0.3\textwidth]{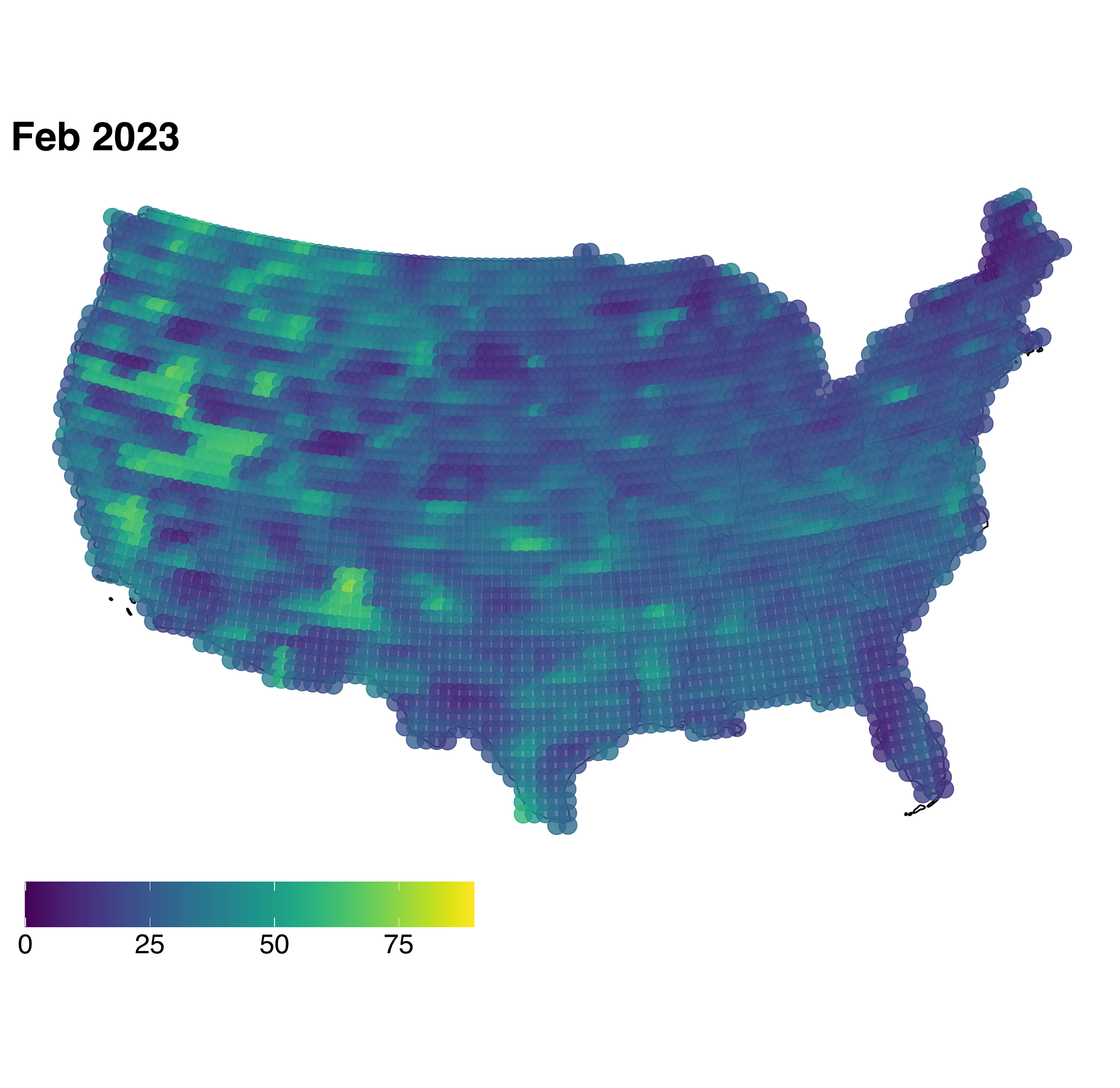} 
     \includegraphics[width=0.3\textwidth]{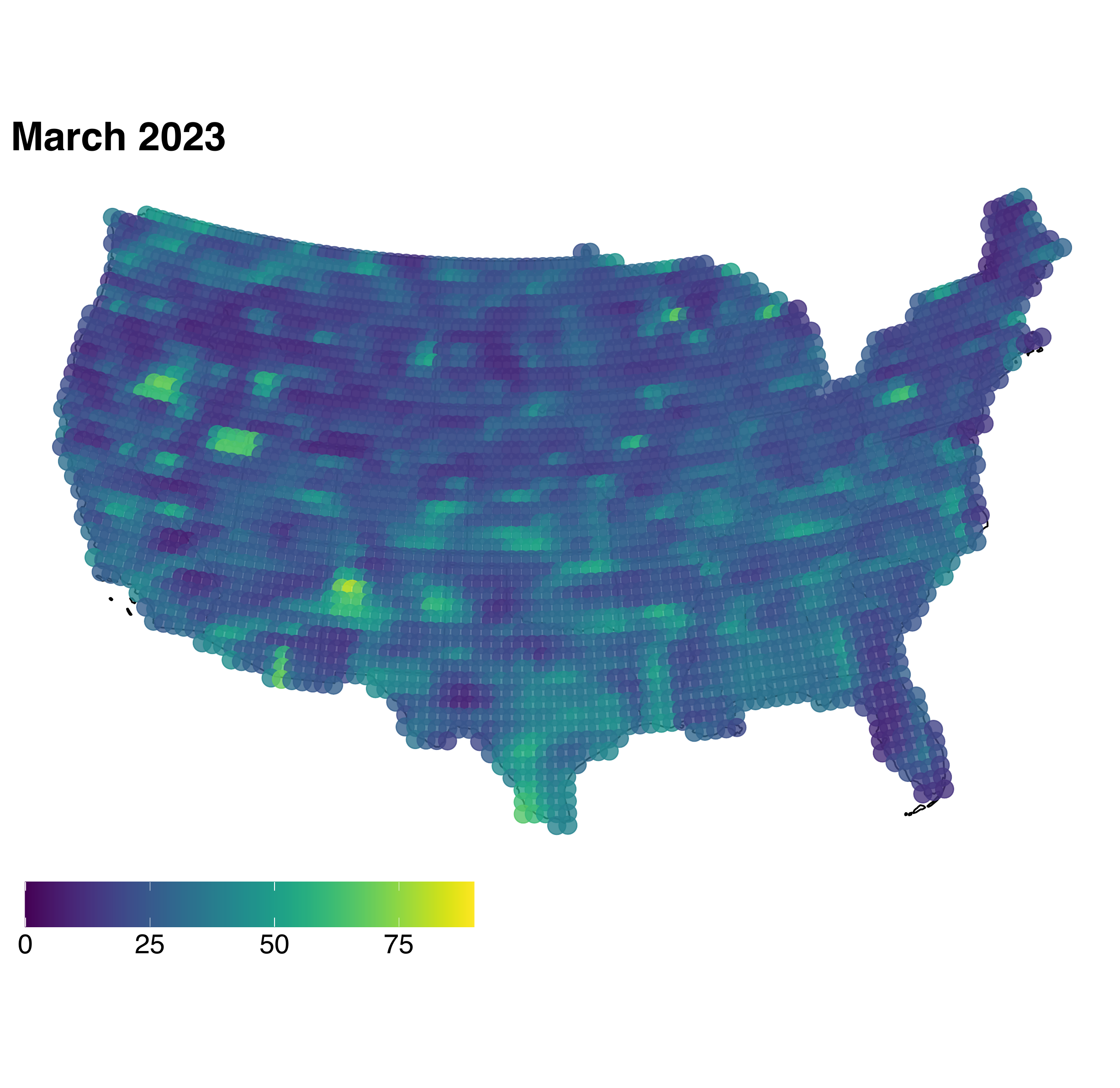} 
      \end{minipage}
    }
    \vspace{-2mm}
    \subfigure[]{
     \begin{minipage}{\linewidth}
     \centering
     \includegraphics[width=0.3\textwidth]{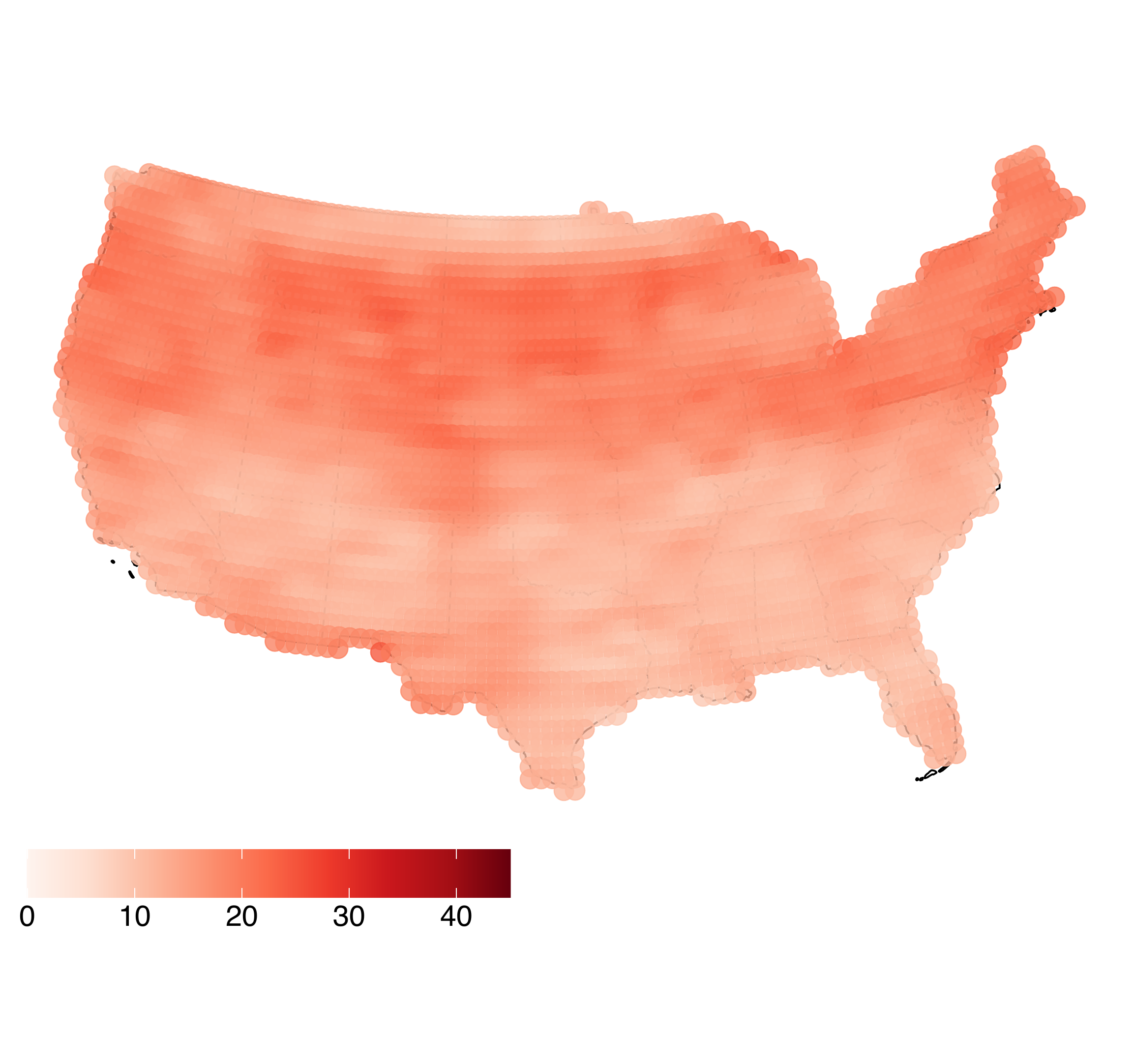}
     \includegraphics[width=0.3\textwidth]{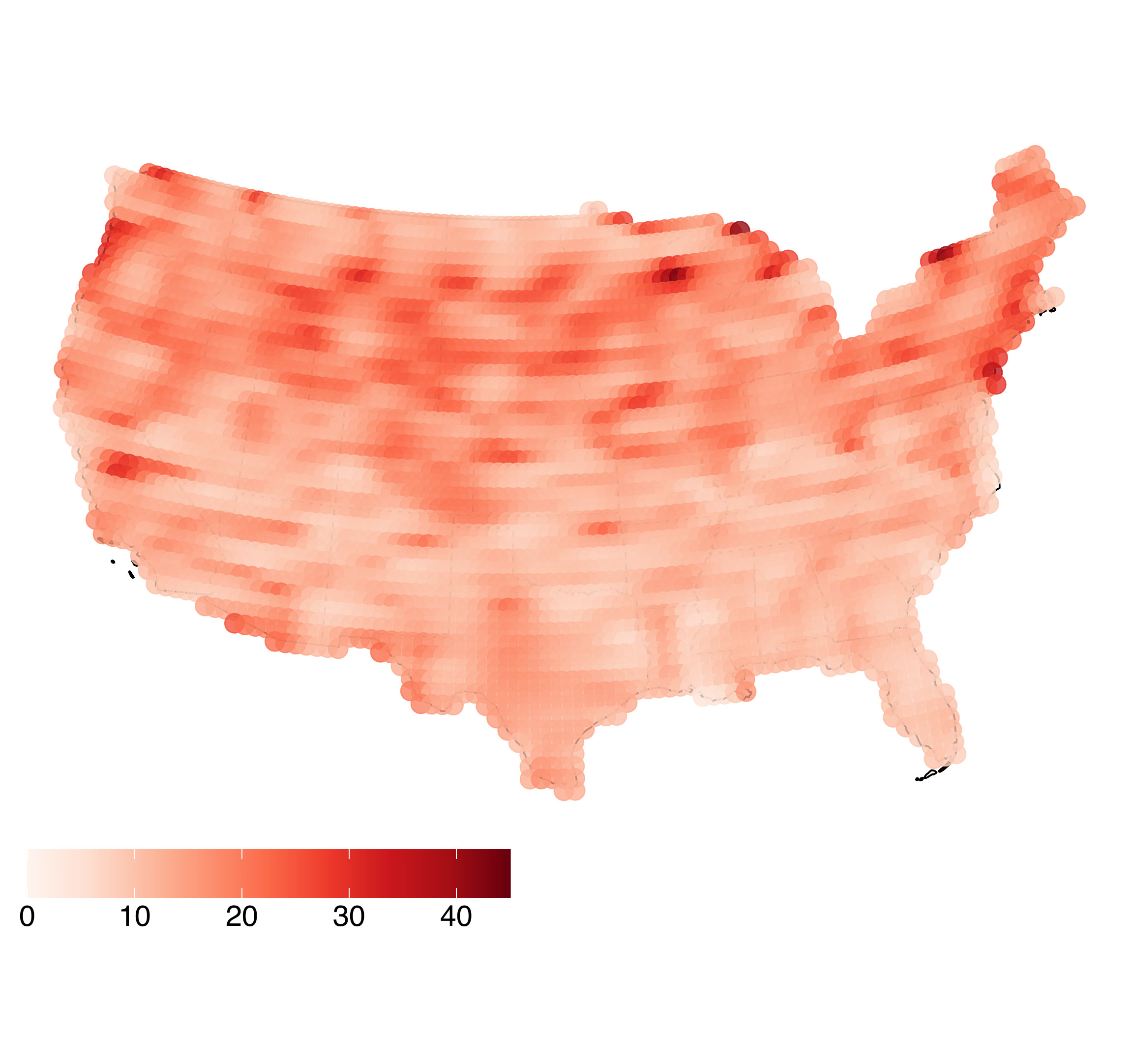}
     \includegraphics[width=0.3\textwidth]{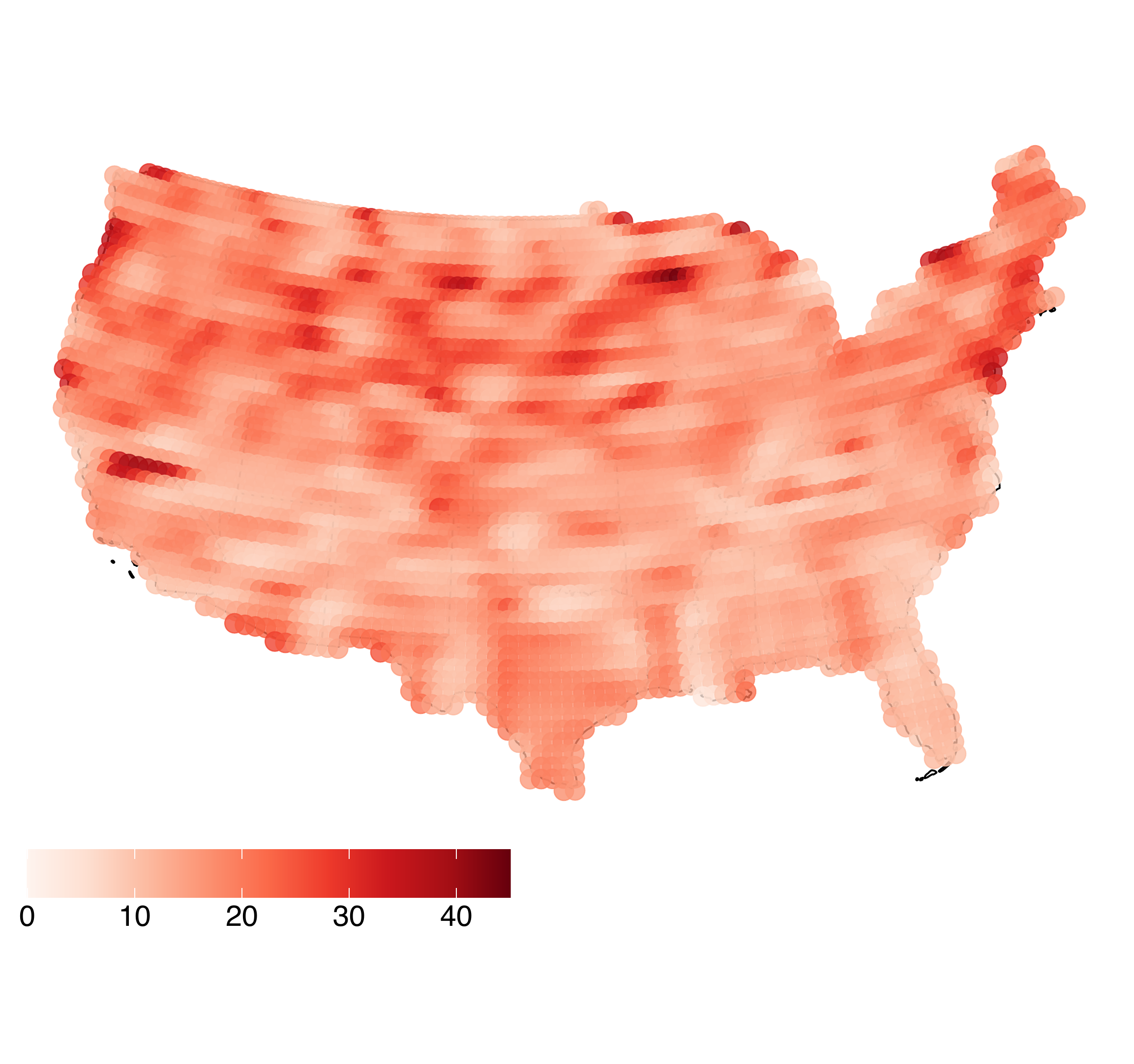}
      \end{minipage}
    }
    
    \caption{(a) Forecast of $PM_{2.5}$ for first three months of 2023 over the United States, and (b) width of the 90\% forecast interval.}
    \label{fig:forecast_future}
\end{figure}

Next, we examine forecasting performance using the last five months as the prediction set. Table~\ref{tab:real_forecast} displays the mean prediction interval width for a 90\% forecast interval as well as the average mean square prediction error for the five time points averaged across all locations. In terms of MSPE, \textbf{QConvLSTM} fared significantly better than alternative modeling frameworks. The forecast for the first three months of 2023 and width of forecast interval are shown in Figure~\ref{fig:forecast_future}. The forecast interval steadily widens with time as expected.

When compared to Figure~\ref{fig:real_interpolation}, the forecasts for future time points are less smooth in space. This is because \textbf{QConvLSTM} and \textbf{QLSTM} forecast at every single site separately. We can alter the proposed architecture and instead offer forecasts for the entire region, but in most cases this significantly degrades model performance due to high dimensionality of the output.

\begin{table}[htb!]
    \centering
    \caption{Average mean square prediction error (MSPE), mean prediction interval width (MPIW) and 90\% interval coverage (COV) for $PM_{2.5}$ forecasts.}
    \vspace{2mm}
        \begin{tabular}{||c | c c c c c||} 
        \hline
        Models & MSPE & SE & MPIW & SE & COV(\%)\\ [0.5ex] 
         \hline\hline
         \textbf{QConvLSTM} & 39.37 & 4.673 & 20.99 & 3.212 & 91.02\\
         \textbf{ARIMA} & 43.27 & 3.677 & 37.89 & 3.956 & 91.15\\
         \textbf{QLSTM} & 43.59 & 6.729 & 19.45 & 3.789 & 90.97\\
          \textbf{GpGp} & 78.56 & 8.374 & - & - & -\\
         \hline
         \end{tabular}
    \label{tab:real_forecast}
\end{table}

\begin{figure}
    \centering
    \includegraphics[width=\textwidth]{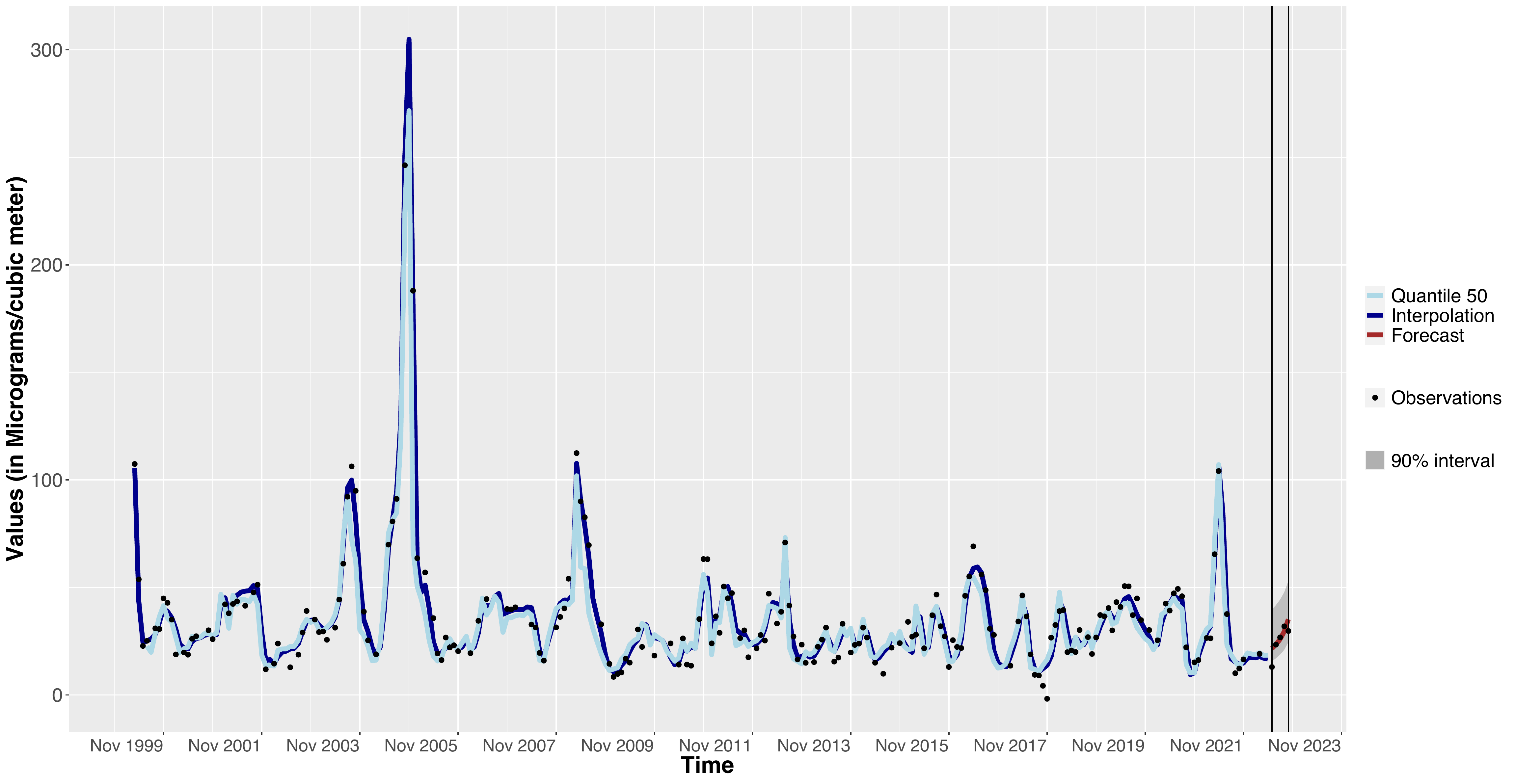} 
    \caption{Forecast using \textbf{QConvLSTM} of $PM_{2.5}$ for one station in San Fransisco ( $37.77^{\circ} N$, $122.42^{\circ} W$).}
    \label{fig:forecast_real_loc1}
\end{figure}

We also examine the probabilistic forecast near San Fransisco ($37.77^{\circ} N$, $122.42^{\circ} W$). The forecasts with 90\% forecast interval are shown in Figure \ref{fig:forecast_real_loc1} for the final five observed months of 2022. Not all time points had observations (black dots), hence interpolation was required. The forecast by \textbf{QConvLSTM} has a high accuracy with reasonable uncertainty intervals.

\section{Computation time}\label{sec:computation_time}

In this section, we compare the computational time of the \textbf{Space-Time.DeepKriging} verses conventional Kriging. DNN uses matrix multiplication over a number of layers to produce its output. In contrast, space-time prediction using Kriging requires an inversion of a $KN \times KN$ matrix. As a result, a single layer $l$ of \textbf{Space-Time.DeepKriging} with $M_{l-1}$ input nodes and $M_{l}$ output nodes will have the time complexity of $O(KNM_{l-1}M_{l})$. The time complexity of \textbf{Space-Time.DeepKriging} with $L$ layers is therefore $O(KN\sum_{l=1}^{L}M_{l-1}M_{l})$. A minibatch \citep{hinton2012neural} is a randomly selected subset of the data used for neural network training. By using minibatch with size $b$, the memory complexity of the training can further be reduced.  Also, the four feed forward networks in each layer of the standard LSTM have set input and output dimensions. Without loss of generality if the input and output dimensions are $A$ and $B$ respectively then $P$ layers will have total time complexity $\mathcal{O}(PbAB)$. Hence the total time complexity for applying \textbf{Space-Time.DeepKriging} and \textbf{QLSTM} simultaniously will be $\mathcal{O}(PbAB + \sum_{l=1}^{L}KNM_{l-1}M_{l})$. On the other hand, time complexity of Kriging with non-separable covariance is $\mathcal{O}(K^3N^3)$. Hence for large $N$ and $K$ our framework with adequate number of layers and nodes is more computationally efficient than traditional Kriging provided $M_l << KN$. 

 The model has been extensively optimized, and in each scenario, the total calculation time has been measured. With one Tesla P100 GPU support, the competition datasets ran in total in an average of \textbf{3.13} minutes. Here we invested in total of 2 days to fine tune the parameters and get the best result in terms of MSPE. However extensive tuning like this may not be necessary in many practical scenarios as the improvements from tuning are very small. The model reported a similar computation time for the second simulation setting. For the $PM_{2.5}$ data with more than 200K locations it took \textbf{7.29} minutes with Tesla P100 GPU and \textbf{19.46} minutes on a standard 16 core CPU to run \textbf{Space-Time.DeepKriging}. 
 
 With GPU capability, running  \textbf{QLSTM} on a single location to extract all the quantiles typically takes 1.37 minutes on average. However, it typically takes \textbf{QConvLSTM}, 12.7 minutes to obtain the median and quantiles for a single location. This is because using a convolution technique on top of an LSTM involves extensive computations. In comparison, training \textbf{GpGp} required \textbf{17.33} minutes on the $PM_{2.5}$ dataset with maximum number of neighbours to be 30. However, when increasing the number of neighbours to 90, the computation time increased drastically to \textbf{1.72} hours and was computationally infeasible with larger neighbourhoods. 
 
 Despite the fact that training \textbf{QConvLSTM} for several locations might take more time, this framework for both \textbf{QLSTM} and \textbf{QConvLSTM} can be  readily implemented on CPU architecture for large datasets without memory concerns. Therefore, compared to any of the conventional statistical methods, these methods are more computationally scalable.

\section{Discussion}\label{sec:conclusion}

In this study, without making any parametric assumptions about the underlying distribution of the data, we have established a novel, easy-to-use methodology for interpolation as well as forecasting for spatio-temporal processes. In order to further enhance our deep learning-based spatio-temporal modeling architecture, we have additionally included nonparametric quantile-based prediction intervals. Our model is compatible with non-stationarity, non-linear relationships and non-Gaussian data. The proposed three methods \textbf{Space-Time.DeepKriging}, \textbf{QLSTM} and \textbf{QConvLSTM} defined here are suitable for large-scale applications. They also require much less computational resources than the traditional statistical techniques.  

In our implementation, we have chosen radial basis functions for spatial embedding and Gaussian kernel for temporal embedding. However, other basis functions such as smoothing spline basis functions \citep{wahba1990spline}, wavelet basis functions \citep{vidakovic2009statistical} as well as different ways of constructing spatio-temporal basis functions such as empirical orthogonal functions (EOFs) can be explored to adapt to certain properties of the spatio-temporal processes. 

 The proposed methods can be further improved by considering different loss functions and better quantifying the uncertainty. \textbf{Space-Time.DeepKriging} uses the simple MSE loss function. Complex terrain behaviors, such as topographical elevation has not been taken into account in our investigation. Utilizing pertinent covariates or developing an adaptive loss function are two ways to include this. Our interval forecasting is conditional on the interpolation from the first step, thus it inherently does not consider the uncertainties for the \textbf{Space-Time.DeepKriging} predictions. Perhaps one way to quantify the two-stage uncertainly is via conditional variance. We can write 

$$
\begin{aligned}
Var(\hat{Y}(\mathbf{s}_0,t_{K+u})) = Var(E(\hat{Y}(\mathbf{s}_0,t_{K+u})|\hat{Y}(\mathbf{s}_0,t_{1}),...,\hat{Y}(\mathbf{s}_0,t_{K}))) + \\
E(Var(\hat{Y}(\mathbf{s}_0,t_{K+u})|\hat{Y}(\mathbf{s}_0,t_{1}),...,\hat{Y}(\mathbf{s}_0,t_{K}))),
\end{aligned}
$$
where $\{\hat{Y}(\mathbf{s}_0,t_{1}),...,\hat{Y}(\mathbf{s}_0,t_{K})\}$ are the interpolations obtained from \textbf{Space-Time.DeepKriging}. These conditional means and variances can be estimated by forecast values obtained by training bootstrap ensembles. Let us take $B_1$ estimates of the interpolation set through bootstraps and ensemble modelling, we call it $T_1 = \{T_{11},...,T_{1B_1}\}$, where $T_{1i} = \{\hat{Y}(\mathbf{s}_0,t_{1})_i,..., \hat{Y}(\mathbf{s}_0,t_{K})_i\}$. For each $T_{1i}$ we train bootstrap ensembles to obtain $B_2$ forecast estimates given as $T_{2i} = \{\hat{Y}(\mathbf{s}_0,t_{K+u})_{i1},...,\hat{Y}(\mathbf{s}_0,t_{K+u})_{iB_2}\}$.  Then

$$
\begin{aligned}
Var(E(\hat{Y}(\mathbf{s}_0,t_{K+u})|\hat{Y}(\mathbf{s}_0,t_{1}),...,\hat{Y}(\mathbf{s}_0,t_{K}))) =
 \frac{1}{B_1}\sum_{i=1}^{B_1} \left(\frac{1}{B_2} \sum_{j = 1}^{B_2} \hat{Y}(\mathbf{s}_0,t_{K+u})_{ij}\right)^2 -& \\
 \left(\frac{1}{B_1}\sum_{i=1}^{B_1}\frac{1}{B_2} \sum_{j = 1}^{B_2} \hat{Y}(\mathbf{s}_0,t_{K+u})_{ij}\right)^2 .
\end{aligned}
$$
Again 

$$
\begin{aligned}
E(Var(\hat{Y}(\mathbf{s}_0,t_{K+u})&|\hat{Y}(\mathbf{s}_0,t_{1}),...,\hat{Y}(\mathbf{s}_0,t_{K}))) = \\
 \frac{1}{B_1}\sum_{i=1}^{B_1} &\left[ \frac{1}{B_2} \sum_{j = 1}^{B_2} (\hat{Y}(\mathbf{s}_0,t_{K+u})_{ij})^2 -
 \left(\frac{1}{B_2} \sum_{j = 1}^{B_2} \hat{Y}(\mathbf{s}_0,t_{K+u})_{ij}\right)^2 \right].
\end{aligned}
$$

Combining these estimates we can obtain the complete uncertainty associated with the 2-stage forecasting procedure. Another interesting avenue of future work for prediction intervals is to extend the conformal prediction methods of \cite{mao2020valid} to this space-time based forecasting. This method is model-free like ours and can be applicable in some more complicated scenarios.

In environmental modeling interpretability can be an important piece. For deep neural network based architectures, one straight forward way of interpretability is to look into the weight matrix of the input layer, which generally gives a basic understanding of the importance of a specific covariate. For example, if the weights corresponding to some covariates are very close to zero, it implies their importance to the model output is minimal. There has been some other recent developments such as Shapley values \citep{merrick} and Local interpretable model-agnostic explanation (LIME) \citep{zafar}. \cite{fuhrman} gives a comprehensive review on explainable AI based on some recent works.

\bibliographystyle{chicago}
\bibliography{reference}

\end{document}